\documentclass[pdflatex,sn-nature]{sn-jnl}

\usepackage{graphicx}%
\usepackage{multirow}%
\usepackage{amsmath,amssymb,amsfonts}%
\usepackage{amsthm}%
\usepackage{mathrsfs}%
\usepackage[title]{appendix}%
\usepackage{xcolor}%
\usepackage{textcomp}%
\usepackage{manyfoot}%
\usepackage{booktabs}%
\usepackage{nameref}%
\usepackage{upgreek}
\usepackage{listings}%
\usepackage{siunitx}
\usepackage{lmodern}
\usepackage{overpic}

\unnumbered

\begin{document}

\title{Global graph features unveiled by unsupervised geometric deep learning}

\author[1]{\fnm{Mirja} \sur{Granfors}}
\author[1]{\fnm{Jesús} \sur{Pineda}}
\author[2]{\fnm{Blanca Zufiria} \sur{Gerbolés}}
\author[2]{\fnm{Joana B.} \sur{Pereira}}
\author*[3, 4]{\fnm{Carlo} \sur{Manzo}}\email{carlo.manzo@uvic.cat}
\author*[1, 5]{\fnm{Giovanni} \sur{Volpe}}\email{giovanni.volpe@physics.gu.se}

\affil[1]{\orgdiv{Department of Physics}, \orgname{University of Gothenburg}, \orgaddress{\street{Origov{\"a}gen 6B}, \city{Gothenburg}, \postcode{SE-41296}, \country{Sweden}}}
\affil[2]{\orgdiv{Department of Clinical Neuroscience}, \orgname{Karolinska Institutet}, \orgaddress{\city{Stockholm}, \country{Sweden}}}
\affil[3]{\orgdiv{Facultat de Ciències, Tecnologia i Enginyeries}, \orgname{Universitat de Vic -- Universitat Central de Catalunya (UVic-UCC)}, \orgaddress{\street{C. de la Laura, 13}, \city{Vic}, \postcode{08500}, \country{Spain}}}
\affil[4]{\orgdiv{Bioinformatics and Bioimaging}, \orgname{Institut de Recerca i Innovaci\'o en Ci\`encies de la Vida i de la Salut a la Catalunya Central (IRIS-CC)}, \orgaddress{\city{Vic}, \postcode{08500}, \state{Barcelona}, \country{Spain}}}
\affil[5]{\orgdiv{Science for Life Laboratory, Department of Physics}, \orgname{University of Gothenburg}, \orgaddress{\street{Origov{\"a}gen 6B}, \city{Gothenburg}, \postcode{SE-41296}, \country{Sweden}}}

\abstract{
Graphs provide a powerful framework for modeling complex systems, but their structural variability poses significant challenges for analysis and classification. To address these challenges, we introduce GAUDI (Graph Autoencoder Uncovering Descriptive Information), a novel unsupervised geometric deep learning framework designed to capture both local details and global structure. GAUDI employs an hourglass architecture with hierarchical pooling and upsampling layers linked through skip connections, which preserve essential connectivity information throughout the encoding–decoding process. Even though identical or highly similar underlying parameters describing a system's state can lead to significant variability in graph realizations, GAUDI consistently maps them into nearby regions of a structured and continuous latent space, effectively disentangling invariant process-level features from stochastic noise. We demonstrate GAUDI's versatility across multiple applications, including small-world networks modeling, characterization of protein assemblies from super-resolution microscopy, analysis of collective motion in the Vicsek model, and identification of age-related changes in brain connectivity. Comparison with related approaches highlights GAUDI's superior performance in analyzing complex graphs, providing new insights into emergent phenomena across diverse scientific domains.}

\maketitle

\section*{\label{sec:Introduction}Introduction}

Complex systems are composed of interrelated elements that interact in varied and often unpredictable ways, leading to emergent properties, dynamic behaviors, and non-linearities that are not apparent when analyzing individual components in isolation~\cite{Argun2021}. 
These systems are generally described by global underlying parameters, offering a global perspective on how they are organized.
However, due to their stochastic nature, even identical underlying parameters can lead to significantly different realizations, where local structures change, while global properties remain statistically indistinguishable. 

Graphs are powerful tools for modeling complex systems~\cite{jalving2019graph}, capturing intricate interactions and dependencies. In these models, nodes typically represent the system's elements while edges denote the relationships between them. The flexibility and scalability of graphs make them particularly valuable for modeling systems of varying sizes and complexities~\cite{holovatch_complex_2017, rathkopf_network_2018}. 
In particular, graphs enable the simultaneous representation of both local and global structures within a complex system, making them invaluable for visualizing, understanding, and reasoning about such systems across various domains.

This graph-based modeling approach helps uncover the underlying parameters of the system, providing deeper insights into its structure and dynamics.
Some complex systems, like Watts-Strogatz small-world graphs, are inherently graph-based~\cite{Watts1998}, while others, such as the human brain, can be readily represented as a graph~\cite{mijalkov_braph_2017}. Additionally, systems that are not inherently in graph form, such as animal flocks and social networks, can be effectively modeled using graphs, with nodes representing the individual agents and the edges capturing their awareness of each other~\cite{bode2011impact}.

In recent years, graph neural networks (GNNs), a class of deep learning models designed to operate on graph-structured data~\cite{battaglia2018relational, velivckovic2023everything}, have gained significant popularity. The inherent structure of GNNs makes them particularly suitable for uncovering hidden patterns and relationships in complex systems. For instance, GNNs utilizing attention mechanisms have been applied to discover hidden interactions in active matter~\cite{ha_unraveling_2021}. Another GNN approach based on message passing has successfully predicted system order parameters in self-driven collective dynamics~\cite{wang_learning_2022}. In addition, a GNN approach integrating message passing with attention mechanisms has been used to estimate dynamical properties in complex biological systems~\cite{pineda_geometric_2023}

In parallel, the performance of GNNs in graph unsupervised learning has been advanced through graph contrastive learning~\cite{velivckovic2018deep}. Graph contrastive learning uses augmentations to create multiple versions of the graph by introducing slight variations but retaining the essential structure and properties. It thus maximizes the similarity between augmented views of the same graph and minimizes the similarity between representations of different graphs. In this way, graph contrastive learning obtains generalizable graph representations for various downstream tasks~\cite{you_graph_2021}.

A major limitation of most existing graph contrastive learning methods is that they derive node representations primarily from local neighborhood information and have inherent limitations in capturing global knowledge, either structural or semantic~\cite{ding2023eliciting}. In addition, their predominant focus on instance-level contrast leads to a latent space where inputs with different augmentations have similar representations, but the underlying semantic structure of the input graph is largely ignored~\cite{li2021prototypical}.

To effectively analyze complex systems, capturing global information and building a well-structured latent space is essential for several reasons. 
First, it enables generalization from observed data to new, unseen scenarios, allowing for accurate interpolation between known states and the prediction of new ones. Second, ensuring that the latent space has physical meaning helps interpreting the system state and understanding the underlying processes.
In addition, for systems that evolve over time, it must ensure that temporal dynamics can be associated with continuous trajectories in the latent space. This capability is fundamental for accurately predicting future states based on current and past observations.

Devising an unsupervised method capable of creating a structured and continuous latent space in which different realizations of the state of a complex system are mapped to nearby points is a challenging task and, to the best of our knowledge, has not been demonstrated in existing approaches.
Such a method must account for the inherent randomness of these systems, where the same global parameters can produce different realizations, resulting in graphs that differ significantly in terms of both connectivity and features. Therefore, the method must reach a level of abstraction that enables pattern recognition at multiple scales and the association of combinations of patterns with a global organization.
This cannot be achieved with graph contrastive learning since structural and semantic augmentations inherently change the state of the system, breaking the continuity and consistency required for a meaningful latent space representation. 

In this article, we propose a Graph Autoencoder Uncovering Descriptive Information (GAUDI), a geometric-deep learning framework that captures multiple essential features and properties of complex systems represented as graphs. By combining graph convolution with MinCut pooling and upsampling, GAUDI gradually reduces the size of a graph to a compressed version in a low-dimensional latent space. Connectivity information is directly passed from the encoder to the decoder so that the latent space can capture the relevant parameters and global structural characteristics of complex systems. In the latent space, realizations of a system with the same underlying parameters are placed close together, assembling a structure that corresponds to the inherent parameter space. The latent space representation is subsequently decoded to reconstruct the input graph, enabling unsupervised training of the model.

We demonstrate GAUDI's capability to uncover descriptive parameters of graph-represented systems across four diverse scenarios: Watts-Strogatz small-world graphs~\cite{Watts1998}, protein assemblies from single-molecule localization microscopy~\cite{lelek2021}, collective behavior in active matter~\cite{Vicsek1995}, and brain connectivity~\cite{cam-can_cambridge_2014}. 
In each case, GAUDI effectively maps system realizations into a structured latent space that reflects underlying physical or biological parameters. By doing so, it facilitates a deeper understanding of the system’s complexity and enables meaningful comparisons between conditions. We benchmarked against established unsupervised graph learning methods, including autoencoders~\cite{ge2021graph} and contrastive learning approaches~\cite{wang_molecular_2022}, GAUDI achieve superior performance in disentangling relevant features and capturing both local and global graph properties.

\section*{\label{sec:Results}Results}
\subsection*{\label{subsec:results_explain_network}GAUDI architecture}

\begin{figure*}[t!]
  \begin{minipage}[c]{0.6\textwidth}
    \includegraphics[width=\textwidth]{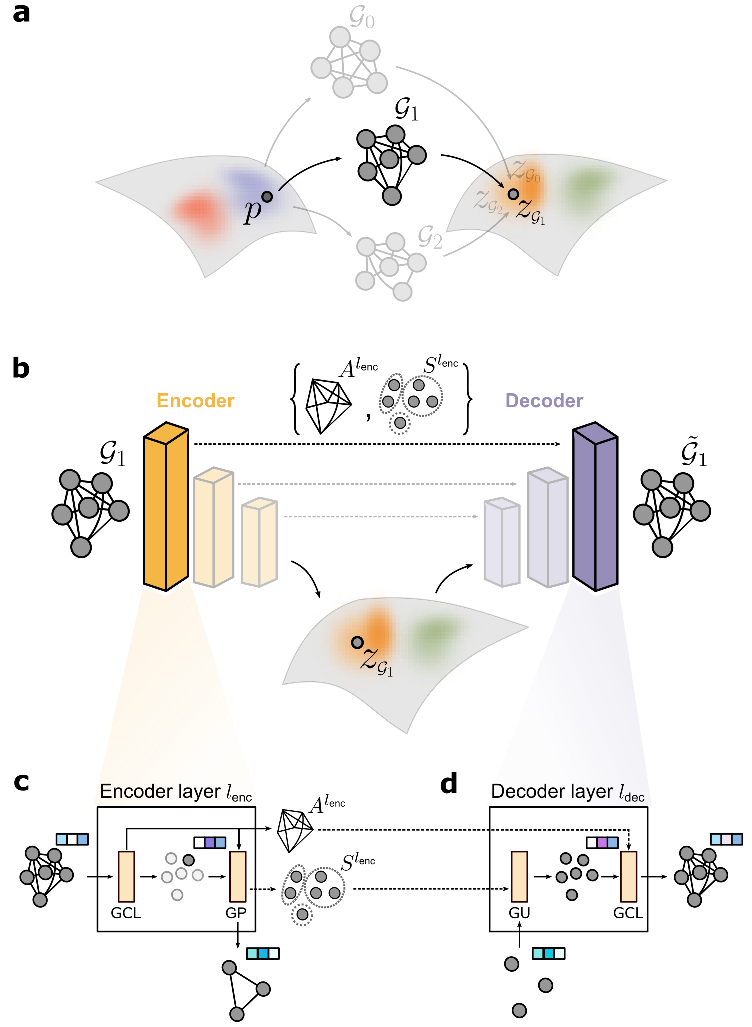}
  \end{minipage}\hfill
  \begin{minipage}[c]{0.42\textwidth}
    \caption{
        \textbf{GAUDI architecture.} \textbf{a},~Graphs with the same underlying parameters $p$ can exhibit significant structural variability. Despite this, GAUDI effectively maps graphs originating from the same parameters close together in the latent space, demonstrating its ability to capture underlying similarities.
        \textbf{b},~GAUDI uses a hierarchical graph-convolutional variational autoencoder architecture, where an encoder progressively compresses the graph into a low-dimensional latent space, and a decoder reconstructs the graph from the latent embedding. At all the levels of compression, adjacency matrices $A^{l_{\text{enc}}}$ and cluster assignment matrices $S^{l_{\text{enc}}}$ are sent directly from the encoder to the decoder to ensure the network embeds the overall structural features.
        \textbf{c},~Each encoder block comprises a Graph Convolution Layer (GCL) followed by a MinCut Graph Pooling layer (GP). The GCL updates the node features of the graph, while the adjacency matrix remains unchanged. The visual separation is only for clarity, illustrating how the adjacency matrix is sent through skip-connections to the decoder block. The GP reduces the graph's dimensionality while retaining essential topological features. The pooling generates a cluster assignment matrix and an adjacency matrix for the pooled graph.
        \textbf{d},~The decoder block mirrors the encoder block structure, featuring a Graph Upsampling (GU) layer to reverse the pooling process and a GCL to accurately reconstruct the graph. At each decoder layer, the cluster assignment matrix from the encoder is used for upsampling, and the adjacency matrix is used in the graph convolution of the reconstructed graph.}
       \label{fig:network}
  \end{minipage}
\end{figure*}

The objective of GAUDI is to map different realizations of the state of a complex system --- described by the same values of the underlying parameters --- into nearby points on the latent space in an unsupervised manner, as schematically shown in Fig.~\ref{fig:network}a.
The realization of the system is represented through a graph, where nodes correspond to its elements and edges denote interactions and relationships.
Each node is characterized by a set of features that reflect the state of the element within the complex system. Graph connectivity is encoded by the adjacency matrix, which indicates which pairs of nodes are connected. 

For other data structures, such as images, this objective is typically achieved through encoder-decoder architectures with an ``hourglass'' structure that first progressively reduces and then progressively increases the dimensionality of the latent space using pooling and upsampling layers. Through self-supervised training aimed at reconstructing the input, these autoencoders learn a low-dimensional representation that captures the essential features of the data~\cite{li2023comprehensive}.

For graphs, autoencoders have primarily been applied for downstream tasks such as link prediction, node classification, and node clustering~\cite{pan_adversarially_2018, guo_multi-scale_2022, wang_attributed_2019}. 
These approaches often do not use pooling and upsampling. Instead, they encode structural information into the nodes and reconstruct the connectivity matrix from the node features, focusing on local features rather than capturing the global structure of the graph.
This choice restricts the network's ability to learn global information about the complex system.
Even recent graph contrastive learning methods designed to preserve global structural and semantic patterns~\cite{ding2023eliciting} often prioritize downstream task performance, without explicitly promoting the formation of a meaningful latent space.

To address these limitations, we propose a graph-convolutional variational autoencoder with an ``hourglass'' structure (Fig.~\ref{fig:network}b).  
At all levels of compression, the encoder explicitly sends the adjacency matrix and the cluster assignment matrix, which encodes the graph’s partitioning during pooling, to the decoder via bridging connections. These connections are similar to the skip connections used in the U-Net~\cite{ronneberger2015u}.
The skip connections help preserve connectivity information, reducing the need for coarser-scale representations to retain local details and thereby encourage the learning of representations that capture the graph's overall structure and high-level features.
The ``hourglass'' architecture is crucial for increasing the perceptive field without inducing oversmoothing~\cite{alon2020bottleneck} and capturing multiscale information, resulting in a parametric representation that retains critical details at the bottleneck.

The encoder consists of a series of identical blocks designed to gradually reduce the size of the graph until only a single node remains, representing the entire graph while preserving critical topological features~\cite{Bianchi2020}. The features of this single node can be viewed as a point in the low-dimensional latent space where complex structures from the original data are encoded. The decoder mirrors the encoder's structure, reconstructing the graph from the latent embedding step by step. 

As depicted in Fig.~\ref{fig:network}c, encoder blocks contain a Graph Convolution Layer (GCL) for the efficient transfer of information between neighboring nodes, followed by a MinCut Graph Pooling layer (GP) that reduces the number of nodes while preserving critical topological features and maintaining key information for reconstruction~\cite{Bianchi2020,grattarola_understanding_2024}. 
Similarly, as shown in Fig.~\ref{fig:network}d, decoder blocks contain a Graph Upsampling layer (GU) to increase the size of the graph and a GCL to allow information to flow between neighboring nodes. 

Through self-supervised training based on graph-to-graph reconstruction, GAUDI learns to align the node features of the reconstructed graph with those of the input graph, thus forcing the latent space to be highly expressive.

\subsection*{\label{subsec:results_watts_strogatz}GAUDI captures the structural parameters of complex systems}

\begin{figure*}
    \centering
    \includegraphics[width=0.95\textwidth]{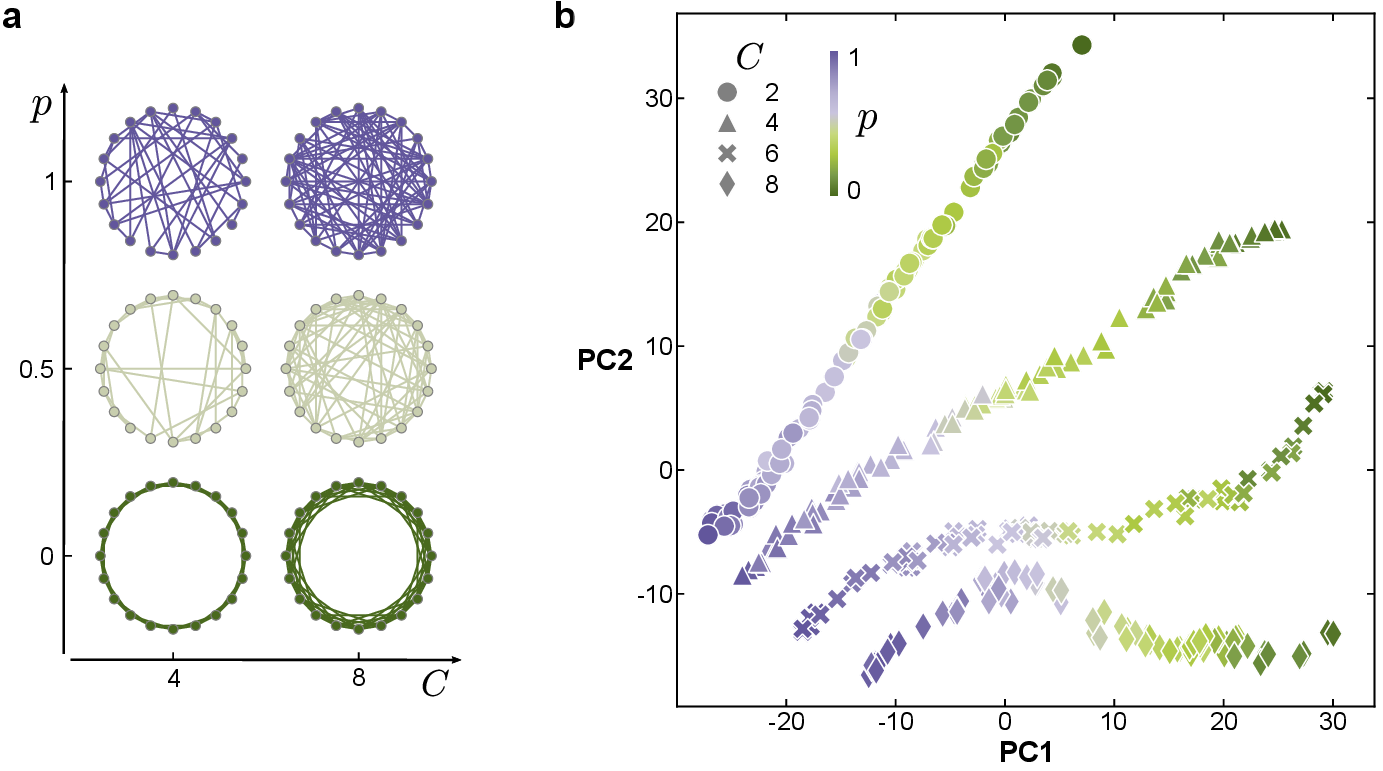}
    \caption{\textbf{GAUDI representation of Watts-Strogatz small-world graphs in latent space}. GAUDI encodes Watts-Strogatz graphs, characterized by node degree $C$ and rewiring probability $p$, into latent variables. \textbf{a}, These graphs model structures ranging from highly ordered to completely random as $p$ increases. As $C$ increases, nodes form more connections, resulting in higher overall connectivity. 
    \textbf{b}, Scatter plot of the first two principal components of the graphs' latent representations, with each point representing a compressed graph. The shape of the scatter points indicates the node degree $C$, where GAUDI effectively clusters graphs based on this parameter. In contrast, the color gradient represents the rewiring probability $p$, showing a smooth transition without distinct clustering, thus illustrating the seamless integration of $p$ variations within the latent space.}
    \label{fig:Watts-Strogatz}
\end{figure*}

To evaluate GAUDI's capability to describe complex systems meaningfully, we first utilized Watts-Strogatz small-world graphs. In these graphs, nodes are arranged in a ring and connected according to the values of two parameters that dictate the structural features of the graph~\cite{Watts1998}. The first parameter, denoted as $C$, represents the number of neighboring nodes to which each node is initially connected, establishing a regular lattice structure. The second parameter, $p$, quantifies the probability of rewiring existing connections, thereby introducing randomness into the network. As $p$ increases from zero to one, the network continuously transitions from a highly ordered structure to a random configuration (Fig.~\ref{fig:Watts-Strogatz}a).

We applied GAUDI to 350 Watts-Strogatz small-world graphs, each comprised of 80 nodes, with various values of $C$ and $p$. Detailed descriptions of the simulations are provided in \nameref{subsec:methods_datasets}. For this dataset, GAUDI's hyperparameters were set to decrease the size of each graph to a single node with eight features at the autoencoder bottleneck.
We used principal components analysis (PCA) to further reduce the dimensionality of the latent space to be able to represent it in two dimensions, as shown in Fig.~\ref{fig:Watts-Strogatz}b. The visualization of the latent space demonstrates that GAUDI distinctly captures the underlying parameters $C$ and $p$ of the Watts-Strogatz graphs, showcasing the neural network's ability to differentiate between graphs based on these parameters.

The discrete parameter $C$, representing the average number of neighboring nodes, produces distinct clusters, whereas the continuous rewiring probability $p$ shows a gradual transition from lower to higher values, indicating a seamless integration of this parameter's variations within the latent space. This demonstrates that the latent space effectively captures both the discrete and continuous nature of the parameters as well as their dynamic ranges.

\subsection*{\label{subsec:supres_results}GAUDI discriminates the morphological features of protein assemblies}

Next, we applied GAUDI to single-molecule localization microscopy (SMLM) data to investigate the morphology of protein assemblies.
SMLM maps molecular organization within biological samples beyond the diffraction limit by using the stochastic blinking of fluorescent emitters~\cite{lelek2021}. 
SMLM image streams are processed to generate point cloud data, where each point corresponds to a localized emitter. 
These point clouds are inherently stochastic due to variations in labeling efficiency, emitter photophysics, and blinking behavior, often resulting in undersampling or oversampling with impact on the estimation of the molecular copy numbers.
Photon noise, along with the use of linkers and antibodies for labeling, further limits localization precision.

Localizations form interconnected networks of relationships at multiple scales. Proximal localizations may represent repeated appearances of the same emitter or multiple emitters labeling the same molecule. Additionally, interactions between molecules can produce hierarchical and organized structures, whose morphology and dimensions are critical to biological function. The collective properties of these localizations reveal emergent behaviors and patterns that cannot be deduced from individual molecules alone, positioning them as a complex system for functional analysis.

To assess GAUDI's ability to accurately map morphological and quantitative features from SMLM point clouds, we used simulated data modeling two structural arrangements, ring-like and spot-like. 
In biological systems, the distinction between ring-like and spot-like structures is crucial for understanding their functional roles and dynamic behaviors. Many essential cellular processes rely on these geometric arrangements to fulfill specific biological tasks.
For instance, nuclear pores form ring-like structures to mediate molecular transport between the nucleus and cytoplasm~\cite{hoelz2011}, while endosomal vesicles display spot-like organization linked to membrane trafficking and recycling~\cite{huotari2011}. Similarly, in mitochondria, rings may correspond to organized cristae arrangements, reflecting the intricate architecture that optimizes ATP production~\cite{mannella2006}. In contrast, spots typically indicate localized zones of activity, such as reactive oxygen species (ROS) production within mitochondria, which plays roles in signaling, stress response, and apoptosis~\cite{murphy2011}.

In addition, cellular structures are not static; they can evolve between spots and rings depending on functional requirements.
The exocyst complex, integral to vesicular trafficking, can appear as spots during initial vesicle tethering to specific membrane sites, later forming rings as the vesicle docks and prepares for fusion~\cite{heider2016, Puig-Tinto2025continuum}.
During cytokinesis, septins transition from dispersed spots to an organized ring that facilitates the formation of the contractile ring, essential for cell division. Conversely, stress responses or signaling events may lead to the dispersion of organized ring structures back into spots, emphasizing the fluid nature of these geometries~\cite{mostowy2012}.

SMLM reliance on stochastic sampling and uneven labeling can result in incomplete reconstructions, where a continuous ring may appear fragmented, mimicking spots, or vice versa. Our simulations incorporated key sources of stochasticity. Each arrangement consisted of a variable number of protein complexes randomly distributed within a defined area. Each complex contained a fixed number of proteins, with a 50\% probability of labeling per protein. To simulate emitter variability and blinking, labeled proteins generated localizations following a geometric distribution. Spatial arrangements accounted for geometric constraints and typical SMLM localization precision. Noise was added by corrupting each point cloud with a random number of uniformly distributed spurious localizations. Additional details are provided in \nameref{subsec:methods_datasets}.

\begin{figure*}
     \centering
     \includegraphics[width=0.98\textwidth]{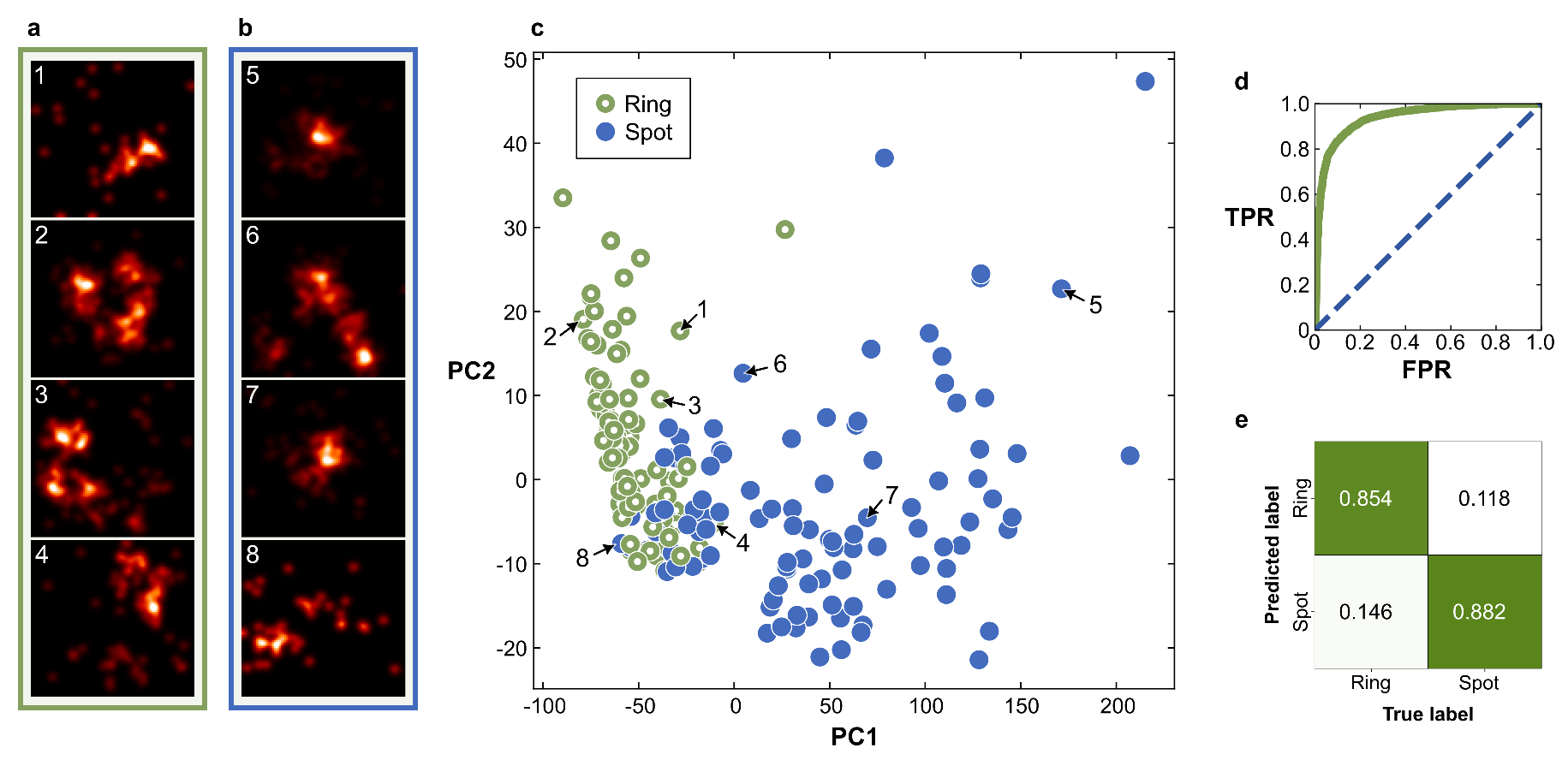}
     \caption{\textbf{Latent space representation of protein assemblies by GAUDI}. GAUDI encodes protein assemblies from simulated single-molecule localization data, categorized into ring-shaped or covering a spot-like area. \textbf{a},~Four examples of the protein assemblies exhibiting a ring-shaped distribution. The molecular localizations are rendered using a Gaussian convolutional kernel. \textbf{b},~Four examples of the protein assemblies belonging to the group generated to follow a spot-like shape. \textbf{c},~Scatter plot showing the first two principal components of the latent representations for 300 samples. Colors denote the distribution type (ring-shaped or spot-like). The arrows indicate the placement in latent space of the samples shown in~\textbf{a} and~\textbf{b}. \textbf{d},~The ROC-curve for the classification using a support vector machine on the latent space for all 10,000 samples~(AUC=0.94). \textbf{e}~Confusion matrix for this classification.}
     \label{fig:superresolution}
\end{figure*}

Illustrative examples of simulated data are shown in Fig.~\ref{fig:superresolution}a,b, where point clouds are rendered as super-resolution images using a 2D Gaussian convolutional kernel.
Due to the inherent stochasticity of SMLM techniques, distinguishing between these two shapes is challenging, as some samples align with the expected structures, while others significantly deviate.
For instance, the images in Fig.~\ref{fig:superresolution}a correspond to ring-like structures, but the shape is difficult to discern in at least three of the four examples (1, 3, and 4). Similarly, Fig.~\ref{fig:superresolution}b displays samples generated to follow a spot-like shape. While examples 5 and 7 closely match this expectation, the remaining examples appear ambiguous and harder to classify accurately.

The structure of localization point clouds lends itself naturally to graph representation, where each localization is a node. To propagate global morphological information, each node was connected to the six nodes furthest from it.
We used GAUDI to analyze these graphs and project the data into an eight-dimensional latent space. The first two dimensions, derived via principal component analysis (PCA), capture the most significant variation in the data, as illustrated in Fig.~\ref{fig:superresolution}c.

GAUDI structures the latent space taking into account the categorical variables that describe the dataset.
The latent vectors form two clear clusters, one primarily associated with ring-shaped arrangements (green markers) and another with spot-shaped arrangements (blue markers).
The separation is further quantified using a support vector machine (SVM), which identifies the optimal decision boundary. A receiver operating characteristic (ROC) curve is generated, with the area under the curve (AUC) measuring the model's classification performance. The ROC curve for the entire dataset (Fig.~\ref{fig:superresolution}d) and the confusion matrix (Fig.~\ref{fig:superresolution}e) show an AUC of 0.94, indicating GAUDI's strong ability to differentiate the two classes.

\subsection*{\label{subsec:vicsek_results}GAUDI characterizes self-driven collective dynamics}

\begin{figure*}
    \centering
    \includegraphics[width = 0.9\textwidth]{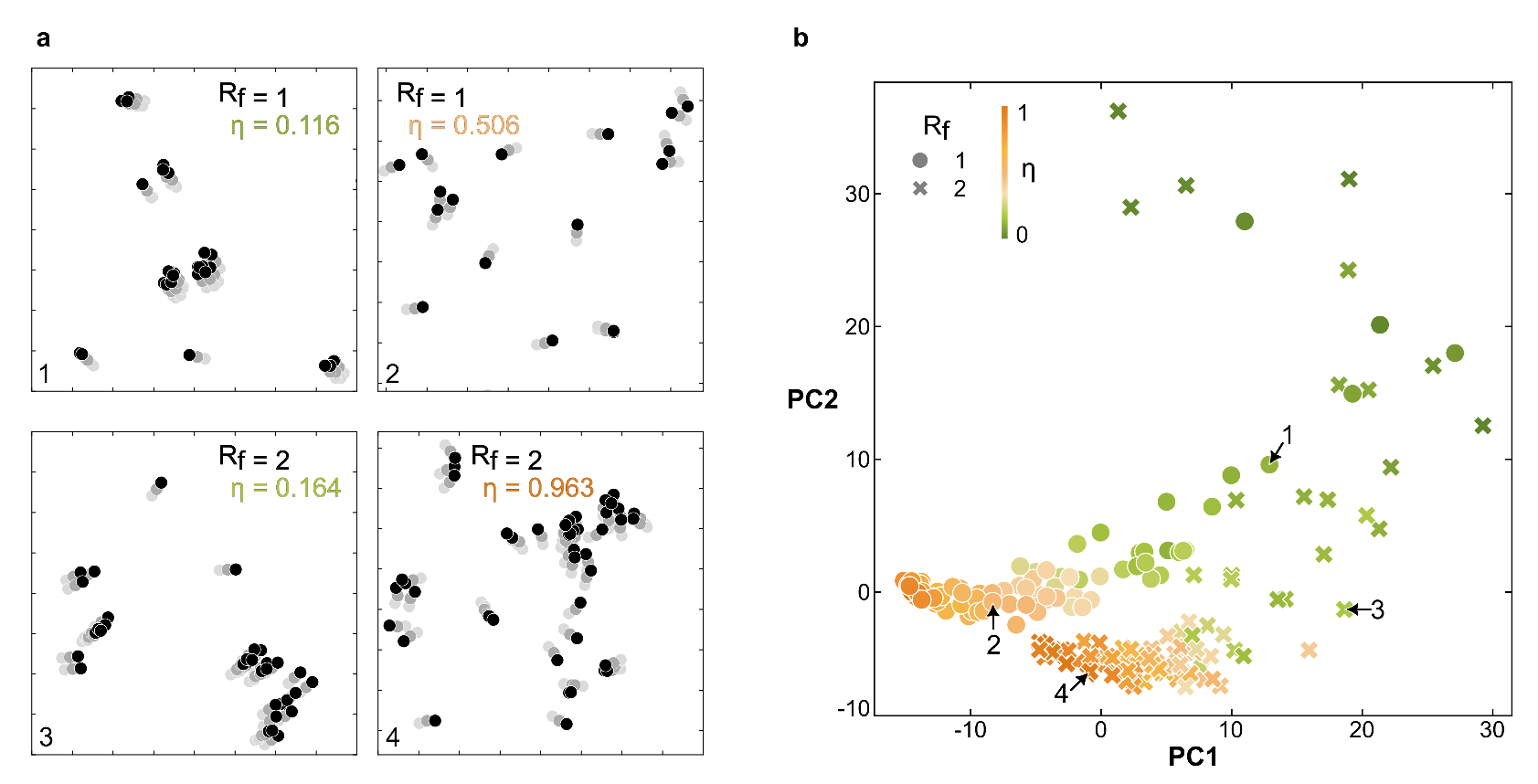}
    \caption{\textbf{GAUDI's latent space representation of self-driven collective behavior}. GAUDI encodes self-driven collective dynamics from Vicsek simulations. \textbf{a,}~Examples of parts of the simulations with varying flocking radius~$R_{\rm f}$ and varying noise level~$\eta$, as specified in the plots. The upper two examples have a smaller flocking radius, resulting in a shorter interaction range, while the bottom two examples have a larger flocking radius. The noise level is lower in the examples to the left, and larger to the right. Three consecutive time steps for each particle are shown, with the darkest color visualizing the last time step. \textbf{b,}~The first two principal components of the latent space of 200 Vicsek model graphs obtained using GAUDI are shown in the scatter plot. The shape of the scatter points depends on the flocking radius $R_{\rm f}$ of the corresponding sample, where GAUDI effectively clusters graphs based on this parameter. The color corresponds to the noise level~$\eta$, showing a smooth transition in the latent space. The arrows indicate the placement in latent space of the samples shown in~\textbf{a}.}
    \label{fig:vicsek}
\end{figure*}

We further evaluated GAUDI's performance on graphs derived from collective dynamics simulated using the Vicsek model~\cite{Vicsek1995}.  
The Vicsek model captures essential features of collective motion observed in nature, such as bird flocking, fish schooling, and insect swarming. These behaviors provide insights into how simple local interactions between individuals can lead to complex group dynamics. In the Vicsek model, particles move at a constant velocity, aligning their direction at each time step with the average direction of other particles within a predefined flocking radius, $R_{\rm f}$. Random noise, scaled by the noise level $\eta$, is added to the updated direction to introduce variability. Further details of the simulations are provided in~\nameref{subsec:methods_datasets}.  
By varying the flocking radius, noise level, and particle density, the particles exhibit transitions from a gas-like phase, where motion is largely independent, to a swarming phase, characterized by coordinated group movement. Examples from these simulations are shown in Fig.~\ref{fig:vicsek}a, illustrating particle positions across three consecutive time frames.  
We conducted simulations with flocking radii of $R_{\rm f} = 1$ and $R_{\rm f} = 2$, and assigned noise levels $\eta$ randomly within the range~$[0, 1]$. Each simulation ran for 8,000~timesteps, but only the last 1,000 were used for analysis to ensure that the system had reached a steady state. For each time frame, we constructed a graph where particles were represented as nodes, connected to their twelve nearest neighbors.

To evaluate GAUDI's performance on graphs derived from collective dynamics, we computed latent space vectors for each time frame and averaged them to represent each simulation. PCA was applied to reduce the dimensionality of the latent space, with the results visualized in Fig.~\ref{fig:vicsek}b. In the scatter plot, marker shapes denote the flocking radius parameter $R_{\rm f}$ and colors represent the noise level $\eta$.
A clear trend dependent on these parameters emerges in the latent space. Simulations with $R_{\rm f} = 1$ form a distinct cluster, well separated from the one created by those corresponding to $R_{\rm f} = 2$. Furthermore, the positions in the latent space correlate with the noise levels: simulations with low noise ($\eta \approx 0$) are concentrated in the upper-right region, whereas those with high noise ($\eta \approx 1$) occupy the lower-left region. This continuous gradient demonstrates that GAUDI captures both the structural differences between flocking radii and the subtler variations introduced by noise.  
By mapping simulations into a well-structured latent space, GAUDI provides a powerful framework for analyzing and differentiating collective behaviors, even under stochastic conditions.

\subsection*{\label{subsec:brain_results}GAUDI predicts aging from anatomical brain scans} 

\begin{figure*}[h!]
    \centering
    \includegraphics[width=\textwidth]{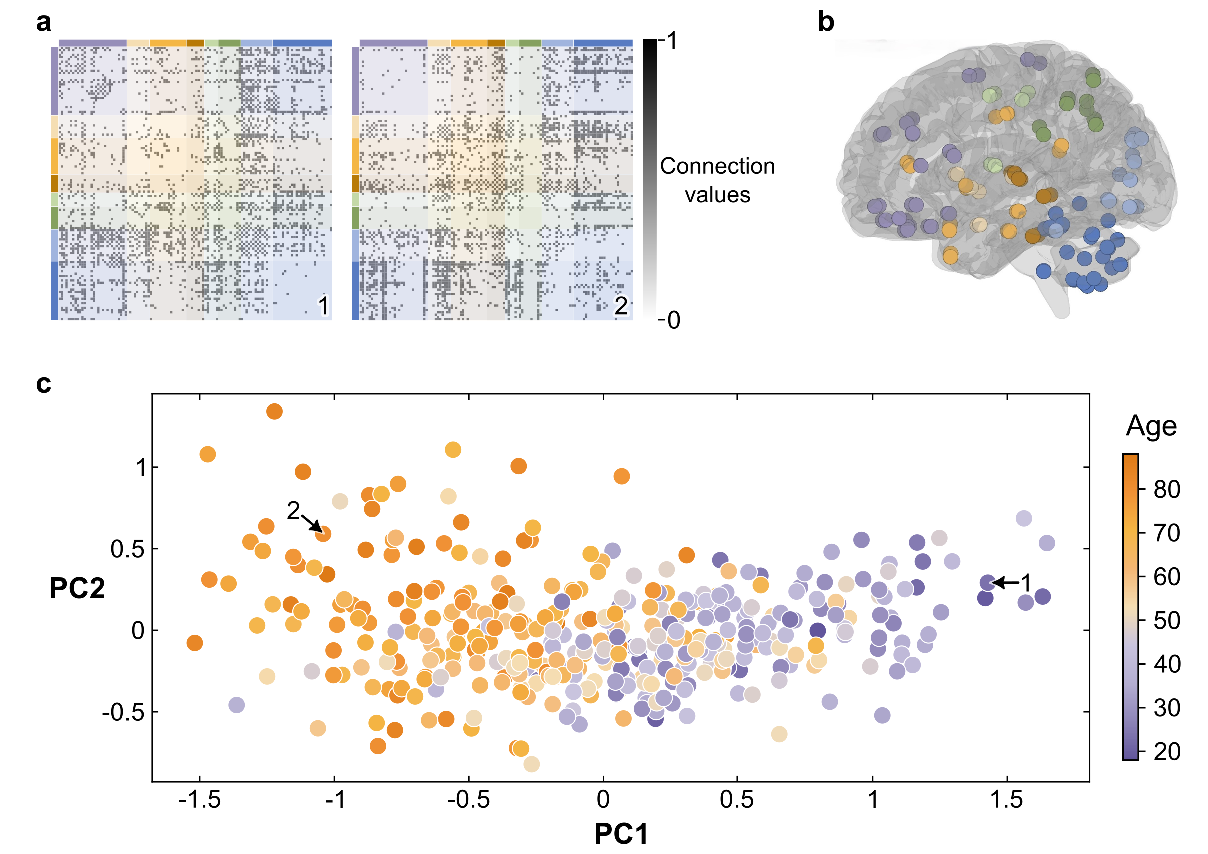}
    \caption{\textbf{GAUDI's latent space captures dependencies between brain graphs and subject ages}. \textbf{a},~Examples of thresholded values of the brain connection graphs of two participants. The graphs are thresholded to only keep the $20\%$ of the connections with highest integrity. The first example is from a 23-year old participant, and the second from a 80-year old. The colors correspond to anatomical groups of brain regions (frontal, parietal, temporal, occipital, subcortical, cerebellar), while the gray-scale indicates the connection integrity. \textbf{b},~Localization of the regions on a brain surface, colored by anatomical group. The illustration is created using BRAPH2~\cite{mijalkov_braph_2017, chang2025braph}. \textbf{c},~The scatter plot shows the first two principal components of the graphs' latent space representations for 400 graphs. The color indicates the age of the participant and the arrows indicate the placement in latent space of the samples shown in~\textbf{a}.}
    \label{fig:brain}
\end{figure*}

Finally, we applied GAUDI to brain diffusion weighted imaging (DWI) scans from the Cambridge Centre for Ageing and Neuroscience (Cam-CAN) dataset~\cite{cam-can_cambridge_2014, taylor_cambridge_2017} to assess its ability to learn meaningful representations of brain connectivity across the lifespan. The dataset includes anatomical connection strengths between the 120 regions of the Automated Anatomical Atlas 2 (AAL2)~\cite{rolls2015implementation} for 633 participants, along with associated demographic information such as age and sex. These regions are treated as nodes of a graph, with edges representing the integrity of the connections between them. To focus on the most significant relationships, we retain only the top 20 percent of the connections with highest integrity values in each sample. In Fig.~\ref{fig:brain}a, the highest connections for two participants are displayed, with the corresponding brain regions shown in Fig.~\ref{fig:brain}b.

GAUDI effectively captures age-related variations in brain connectivity patterns. The first two principal components of GAUDI's eight-dimensional latent space for the CamCAN graphs are shown in Fig.~\ref{fig:brain}c, with the colors corresponding to the participants' ages. The positioning of the graphs in the latent space reveals a clear age-dependent trend. Graphs from younger participants tend to cluster in the lower-right corner of the plot, while graphs from older participants are located higher and slightly to the right. This trend indicates that GAUDI can capture age-related differences in brain connectivity, highlighting its potential to provide valuable insights into how brain connectivity changes across the lifespan.

To further evaluate the quality of the latent space, we calculated the coefficient of determination $R^2$, which quantifies how much the independent variable (the latent space) explains the variation in the dependent variable (age). This metric represents the proportion of variance in the latent space that can be attributed to age, providing insight into the strength of the correlation between age and the learned representation~\cite{wright1921correlation, chicco2021coefficient}. A linear regression analysis of age against the latent space produced a $R^2$ value of 0.54, suggesting a moderate correlation between the latent space and the participants' ages.
Additionally, we applied SVM to classify participants into two age groups: those younger than the median age of 55 years and those 55 or older. We assessed the classification performance using a ROC curve and calculated the AUC. The model achieved an AUC of 0.87, highlighting GAUDI's strong ability to differentiate participants by age within the latent space. This robust discrimination reinforces the potential of GAUDI for advancing age-related brain research.

\subsection*{GAUDI outperforms existing methods across diverse tasks \label{subsec:comparison}}

To benchmark GAUDI against state-of-the-art methods for unsupervised graph representation learning, we compare its performance with two alternative approaches: a graph contrastive learning approach~\cite{wang_molecular_2022} and a graph autoencoder~\cite{ge2021graph}.

We selected these benchmarks as they represent prominent and conceptually distinct approaches in unsupervised graph learning. Graph contrastive learning with structural augmentations is a widely used strategy for learning graph-level representations, while autoencoders have recently been designed to capture both local node-level and global graph features. Moreover, these methods generate a single latent representation per graph, making them directly comparable to GAUDI without requiring substantial architectural modifications. We evaluate all models consistently across the four graph-represented systems studied in this work.

\begin{figure}
  \centering
  \hspace{1.25cm}
  \begin{overpic}[width=0.85\textwidth]{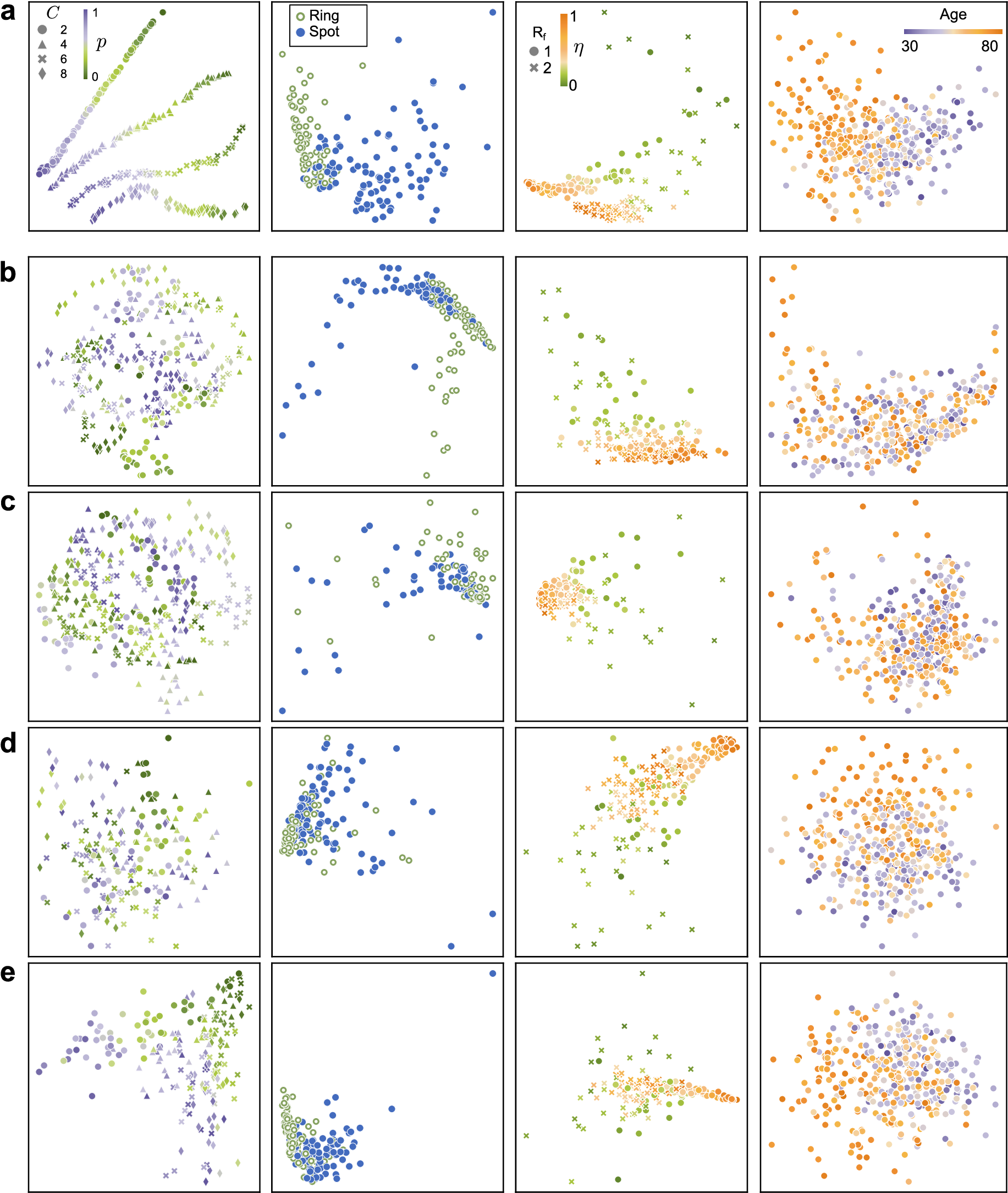}
    \put(-12,89){\small{GAUDI}}
    \put(-12,68){\small{CL-n}}
    \put(-12,48.5){\small{CL-s}}
    \put(-12,28){\small{MIAGAE8}}
    \put(-12,8){\small{MIAGAE64}}

    \put(3,101.7){\small{Watts-Strogatz}}
    \put(28.5,101.7){\small{SMLM}}
    \put(49,101.7){\small{Vicsek}}
    \put(63.5,101.7){\small{Brain connectivity}}
  \end{overpic}
  \caption{\textbf{Comparison of latent space representations obtained with different methods.} The first two principal components of the latent representations are shown for each method (rows) and dataset (columns): (a) GAUDI reproduced from Figs.~\ref{fig:Watts-Strogatz}b, \ref{fig:superresolution}c, \ref{fig:vicsek}b, and \ref{fig:brain}c; (b) graph contrastive learning with node and edge masking (CL-n); (c) contrastive learning with subgraph removal (CL-s); (d) MIAGAE with 8 latent dimensions (MIAGAE8); and (e) MIAGAE with 64 latent dimensions (MIAGAE64). Each column corresponds to one of the datasets analyzed in this study: Watts–Strogatz, SMLM, Vicsek, and brain connectivity graphs. Each point represents a graph, with marker shape and color indicating its generating parameters.}
  \label{fig:summary}
\end{figure}

We begin with a qualitative analysis by visualizing the first two principal components of the latent representations, as shown in Fig.~\ref{fig:summary}.
Across all examples, GAUDI effectively captures both local and global aspects of the graphs’ underlying parameters (Fig.~\ref{fig:summary}a). Locally, it places similar graphs near each other, forming tight clusters based on graph characteristics. Globally, it organizes the latent space in a structured and interpretable way, such that the graph-generating parameters are reflected in the global layout of the embeddings.

For contrastive learning, we implemented the method by Wang et al.~\cite{wang_molecular_2022}, which augments graphs by applying node and edge masking or subgraph removal, and optimizes a contrastive loss that brings augmentations of the same input graph closer in latent space. To adapt the method to our data, we introduced minor architectural and hyperparameter modifications. Specifically, we replaced the input embedding layer with a linear layer to accommodate our node and edge features, and set the latent dimension to 8 to ensure fair comparison. We evaluated two variants: node and edge masking (CL-n, 25\% masked, Fig.~\ref{fig:summary}b) and subgraph removal (CL-s, 25\% removed nodes and edges, Fig.~\ref{fig:summary}c). Both CL-n and CL-s group similar graphs locally, but fail to maintain a meaningful global organization in the latent space. This results in clusters of similar graphs, but with dissimilar graphs also frequently embedded nearby, making it difficult to infer parameters from the representation.

To compare GAUDI with a graph autoencoder, we used the Multi-kernel Inductive Attention Graph Autoencoder (MIAGAE) by Ge et al.~\cite{ge2021graph}. MIAGAE reduces graph size via multi-kernel inductive convolutions and attention-based pooling in the encoder, and reconstructs it through inductive upsampling layers. It is trained using the mean squared error between the original and reconstructed node features. Since MIAGAE does not aggregate graphs to a single node, we average the encoded node features to obtain a single latent vector per graph. We tested both a reduced version (MIAGAE-8, Fig.~\ref{fig:summary}d) constrained to a 8-dimensional latent space for comparability, and the default version (MIAGAE-64, Fig.~\ref{fig:summary}e) with 64 dimensions.
Both MIAGAE variants preserve some local structure but lack a consistent global organization. While some similar graphs are placed together, marker shapes indicating categorical variables are intermixed, and color gradients reflecting continuous parameters lack a coherent trend. This limits the interpretability of the embeddings and impairs the extraction of generative parameters.

\begin{table}
\centering
\footnotesize
\renewcommand{\arraystretch}{1.2}
\begin{tabular}{l@{\hskip 12pt}l@{\hskip 8pt}c@{\hskip 5pt}c@{\hskip 1pt}c@{\hskip 10pt}c}
\toprule
Dataset & Method & 
 \parbox[c][2.5em][c]{1.5cm}{\centering 1-NN \\ accuracy} 
 & \parbox[c][2.5em][c]{2.5cm}{\centering Isomap gradient \\ continuity} 
 & \parbox[c][2.5em][c]{1.5cm}{\centering AUC} 
 & \parbox[c][2.5em][c]{1.5cm}{\centering $R^2$} \\
\midrule
\multirow{5}{*}{Watts-Strogatz}
 & GAUDI            & \textbf{1.00} & \textbf{0.99} &  -   &  -  \\
 & CL-n             &     0.51      &     0.24      &  -   &  -  \\
 & CL-s             &     0.61      &     0.22      &  -   &  -  \\
 & MIAGAE8          &     0.37      &     0.28      &  -   &  -  \\
 & MIAGAE64         &     0.50      &     0.77      &  -   &  -  \\
\midrule
\multirow{5}{*}{SMLM} 
 & GAUDI            & \textbf{0.80} &  -  & \textbf{0.92} (\textbf{0.94}*) &  -  \\
 & CL-n             &     0.78      &  -  &     0.84 (0.86*)      &  -  \\
 & CL-s             &     0.61      &  -  &     0.61 (0.76*)      &  -  \\
 & MIAGAE8          &     0.65      &  -  &     0.83 (0.86*)      &  -  \\
 & MIAGAE64         &     0.75      &  -  &     0.90              &  -  \\
\midrule
\multirow{5}{*}{Vicsek} 
 & GAUDI            & \textbf{0.93} & \textbf{0.98} &  -  &  -  \\
 & CL-n             &     0.49      &     0.88      &  -  &  -  \\
 & CL-s             &     0.75      &     0.95      &  -  &  -  \\
 & MIAGAE8          &     0.80      &     0.82      &  -  &  -  \\
 & MIAGAE64         &     0.79      &     0.82      &  -  &  -  \\
\midrule
\multirow{5}{*}{\shortstack[l]{Brain \\ connectivity}}
 & GAUDI            &  -  & \textbf{0.70} & \textbf{0.87} (\textbf{0.87}*) & \textbf{0.51} (\textbf{0.54}*) \\
 & CL-n             &  -  &      0.14     &     0.56 (0.63*)      &      0.04 (0.12*)     \\
 & CL-s             &  -  &      0.16     &     0.58 (0.65*)      &      0.05 (0.08*)     \\
 & MIAGAE8          &  -  &      0.08     &     0.74 (0.74*)      &      0.27 (0.31*)     \\
 & MIAGAE64         &  -  &      0.52     &     0.78              &      0.36             \\
\bottomrule
\end{tabular}
\caption{\textbf{Quantitative comparison of GAUDI, graph contrastive learning, and MIAGAE.}  
We report the 1-Nearest Neighbor (1-NN) accuracy, class-wise Isomap gradient continuity score, area under the ROC curve (AUC), and coefficient of determination ($R^2$) for each method on four datasets: Watts–Strogatz small-world graphs, single-molecule localization microscopy (SMLM), Vicsek model simulations, and brain connectivity graphs from the CamCAN dataset. Note that not all metrics are applicable to all datasets, as their computation depends on the presence of discrete labels or continuous generative parameters. Contrastive learning is evaluated with node and edge masking (CL-n) and subgraph removal (CL-s). MIAGAE is tested with 8 (MIAGAE8) and 64 (MIAGAE64) latent dimensions. All metrics are computed from the first two principal components (PC1 and PC2) of the latent space, except where noted (*), in which case all 8 latent dimensions are used. Higher scores indicate better performance. Bold values indicate the best performance for each metric within each dataset.}
\label{tab:comparing_methods}
\end{table}

To perform a quantitative comparison of the methods' performance, we computed several metrics, summarized in Table~\ref{tab:comparing_methods}. First, we use the 1-Nearest Neighbor (1-NN) accuracy in the PCA plane to measure whether graphs from the same class form local clusters while remaining distinct from others. Next, to assess whether continuous parameters vary smoothly within classes, we compute a class-wise Isomap gradient continuity score~\cite{Tenenbaum_global_2000}, defined as the mean absolute Spearman rank-order correlation~\cite{spearman_proof_1904, zwillinger_standard_2000} between the Isomap coordinates and the continuous parameter of interest. For binary classification tasks, we report the area under the ROC curve (AUC) from a support vector machine trained on the latent space. Finally, to quantify how well the latent representations explain variation in continuous variables, we compute the coefficient of determination ($R^2$). 
Note that some metrics are dataset-specific, as their applicability depends on the presence of either continuous generative parameters or discrete class labels.
Full metric details are provided in \nameref{subsec:metrics}.

As shown in Table~\ref{tab:comparing_methods}, GAUDI consistently outperforms both graph contrastive learning and MIAGAE across all metrics and datasets. In particular, it achieves near-perfect class-wise Isomap gradient continuity scores for the Watts–Strogatz (0.99) and Vicsek (0.98) graphs, indicating an almost monotonic embedding of the continuous parameters $p$ and $\eta$ within the respective classes. For the SMLM data, GAUDI achieves the highest AUC (0.92 using the first two principal components and 0.94 when using all 8 latent dimensions), outperforming all other methods in this binary classification task. GAUDI also yields the highest continuity score (0.70) on the brain connectivity data and the highest $R^2$ value (0.51), further demonstrating its ability to uncover structure in more complex and noisy biological graphs. This strong global organization, combined with accurate local clustering, highlights GAUDI’s ability to simultaneously capture discrete structure and continuous variation—something that neither contrastive learning nor MIAGAE achieve effectively across all datasets.

\section*{\label{sec:Discussion}Discussion}

GAUDI represents a significant advancement in the unsupervised analysis of complex systems that can be represented as graphs. It excels at learning compact, lower-dimensional latent space representations that distill essential graph features, enabling a more interpretable understanding of their structure and dynamics.

A defining feature of GAUDI is its hierarchical encoder-decoder architecture, which applies graph convolutional layers at multiple scales with pooling operations that progressively reduce the size of the graph. This design enables information to be propagated both locally and globally across the graph while being compressed into a latent representation. Skip connections between the encoder and decoder complement this hierarchical design by preserving connectivity information.
By integrating information across scales, GAUDI's architecture reduces the reliance on local connectivity patterns alone and encourages the latent space to capture higher-order graph features, such as node attribute interactions, local neighborhood roles, and broader structural properties like clusters or communities. As a result, GAUDI learns compact graph representations that reflect essential graph topology while remaining robust to variability introduced by stochastic processes.

GAUDI’s emphasis on process-level invariance distinguishes it from conventional graph contrastive learning methods. While traditional graph contrastive learning approaches prioritize instance-level transformations, treating augmented views as variations of the same input, GAUDI targets invariance across realizations of the same stochastic process. This shift enables it to abstract away random artifacts and instead encode invariant characteristics of the underlying generative process. Consequently, GAUDI’s latent representations not only preserve the local structure of graphs but also maintain global consistency in the latent space, making it particularly suited for applications involving stochastic dynamics and complex systems, where variability stems from intrinsic randomness rather than deterministic augmentations.

Compared to state-of-the-art methods such as graph contrastive learning and MIAGAE, GAUDI consistently achieves superior performance across a range of datasets and metrics. Unlike contrastive learning approaches and other autoencoders, which often fail to organize the latent space globally --- resulting in poor interpretability of continuous parameters --- GAUDI produces structured and disentangled latent spaces that reflect both categorical and continuous generative variables. GAUDI's ability to preserve monotonic trends in continuous parameters and maintain separability across classes demonstrates its effectiveness in encoding both local and global structure, even in the presence of noise or intrinsic stochasticity.

One of GAUDI’s most significant strengths is its ability to extract meaningful features across diverse datasets. This versatility positions it as a powerful tool for analyzing datasets with varying levels of complexity, noise, and stochasticity. However, its performance could be influenced by dataset-specific factors, such as graph size, sparsity, and noise levels. Sparse graphs may challenge the encoder’s ability to pool meaningful information, while dense graphs could overwhelm the model with redundant features. These challenges highlight the importance of tailoring GAUDI’s architecture and hyperparameters to the nuances of each dataset.

Despite its strengths, GAUDI faces challenges with incomplete class separation in some datasets. Factors such as overlapping feature distributions, insufficient training variability, or limitations inherent to unsupervised learning may contribute to this issue. Addressing these limitations may require incorporating task-specific loss functions, leveraging semi-supervised learning, or introducing additional regularization techniques to enhance separation in the latent space.

GAUDI’s ability to generalize across datasets with varying structures and scales is both a strength and a challenge. Graphs of vastly different sizes or topological patterns may require flexible pooling strategies and normalization techniques to ensure consistent representation. For example, mapping both small and large graphs to latent spaces of comparable scale without losing essential details will necessitate further architectural refinements.

Future research could expand GAUDI's utility by leveraging $\beta$-variational autoencoders to determine the minimal meaningful dimensionality of the latent space~\cite{higgins2017beta}. This approach, recently proven effective for analyzing stochastic datasets~\cite{fernandez2024learning}, could enhance GAUDI's capacity to disentangle key features from noisy inputs while maintaining a minimal and interpretable latent space.

GAUDI’s architecture could also be extended to other domains, such as dynamic biological processes or temporal networks, enabling the analysis of evolving systems like protein interactions or traffic patterns. Additionally, adapting GAUDI to alternative graph representations—such as heterogeneous graphs, hypergraphs, or multiplex networks—could broaden its applicability to multi-modal datasets and capture more complex relationships between elements.

\section*{\label{sec:Methods}Methods}

\subsection*{\label{subsec:methods_network}Description of GAUDI}

GAUDI architecture follows a U-shape scheme featuring encoder and decoder pathways (Fig.~\ref{fig:network}b). The encoder receives the input graph and compresses it to a single latent representation that captures essential global structure and features. 
The decoder, in turn, mirrors the encoder and aims to reconstruct the graph by leveraging the latent representation to generate an output that reflects the learned global properties.
The graph connectivity information, including both adjacency matrices and clustering matrices, is transmitted through skip connections directly from the encoder to the decoder. This design helps the decoder preserve essential local connectivity details for graph reconstruction while allowing GAUDI´s latent representation to prioritize capturing the global properties of the graph, ensuring a balanced representation that emphasizes overarching structural patterns without overfocusing on localized intricacies.

Prior to the encoder, the input graph undergoes a preprocessing stage consisting of two message-passing layers. Each message passing layer equips the nodes with information of their topological neighborhood, i.e.,
\begin{equation}
\mathbf{m}_{ij} = \gamma\left( \left[ \mathbf{v}_i, \mathbf{v}_j, \mathbf{e}_{ij} \right] \right),
\end{equation}
where $\mathbf{m}_{ij}$ denotes the message passed from node $j$ to node $i$, $\left[ , \right]$ represents concatenation, $\mathbf{v}_i$ and $\mathbf{v}_j$ are the feature vectors of nodes $i$ and $j$, respectively, $\mathbf{e}_{ij}$ denotes the edge features of the edge between nodes $i$ and $j$, and the function $\gamma$ is a fully-connected dense layer followed by a ReLU activation. Subsequently, the features of each node are updated by aggregating the messages $\mathbf{m}_{ij}$ received from its neighbors, as follows,
\begin{equation}
\tilde{\mathbf{v}}_i = \phi \left( \left[ \mathbf{v}_i, \sum_{j \in \mathcal{N}_i} \mathbf{m}_{ij} \right] \right),
\end{equation}
where $\mathcal{N}_i$ represents the neighborhood of node $i$, and $\phi$ is implemented as a fully-connected dense layer followed by a ReLU activation.

The encoder architecture is composed of three hierarchical building blocks, each integrating a Graph Convolution Layer (GCL) and a Graph Pooling (GP) operation (Fig.~\ref{fig:network}c).
At each encoder block level \(l\), the GCL updates the node features, as follows, 
\begin{equation}
\mathbf{v}_{\text{conv}, i}^{l} = \delta^{l} \left( \left[ \mathbf{v}_{\text{enc}, i}^{l}, \sum_{j \in \mathcal{N}_i} \psi^{l}\left(\mathbf{v}_{\text{enc}, j}^{l}\right) \right] \right),
\label{eq:GCL}
\end{equation}
where \(\mathbf{v}_{\text{enc}, i}^{l}\) represents the features of node \(i\) at the encoder level $l$. Accordingly, in the first encoder block, \(\mathbf{v}_{\text{enc}, i}^{l} = \tilde{\mathbf{v}}_i\). The function $\psi^{l}$ processes the features of the neighboring nodes $j \in \mathcal{N}_i$, and the resulting outputs are aggregated.
 On the other hand, $\delta^{l}$ transforms the concatenated features, which include the current node's value and the aggregated information, allowing the GCL to capture complex relationships and dependencies within the graph. Both $\psi^{l}$ and $\delta^{l}$ are implemented as fully connected dense layers followed by a ReLU activation function.

Following the GCL, the GP operation relies on MinCut pooling to learn progressively coarser representations of the input graph. 
MinCut pooling is a trainable graph pooling technique that clusters nodes based on their feature similarities and the strength of their connections ~\cite{Bianchi2020}.
The pooling process is guided by a cluster assignment matrix $\mathbf{S}$. This matrix is learned using a two-layer, fully-connected neural network. The first layer utilizes a ReLU activation function, while the second layer employs a softmax activation to ensure that the rows of $\mathbf{S}$ sum up to unity.
Finally, the node features of the pooled graph are derived from $\textbf{S}$ as,
\begin{equation}
\mathbf{V}_{\text{enc}}^{l + 1} = (\mathbf{S}^l)^\top \mathbf{V}_{\text{conv}}^l,
\end{equation}
where $\mathbf{V}_{\text{conv}}^{l} = [\mathbf{v}_{\text{conv}, 1}^{l}; \mathbf{v}_{\text{conv}, 2}^{l}; \dots; \mathbf{v}_{\text{conv}, N_v^{l}}^{l}]$ is the node feature matrix, with \(N_v^{l}\) representing the total number of nodes in the graph at level $l$. Similarly, the graph connectivity, represented by the adjacency matrix $\mathbf{A}^{l+1} \in \mathbb{R}^{N_v^{l+1} \times N_v^{l+1}}$, is estimated for the pooled graph using the cluster assignment matrix, as follows,
\begin{equation} 
\hat{\mathbf{A}}^{l+1} = (\mathbf{S}^l)^\top \mathbf{A}^{l} \mathbf{S}^l, 
\end{equation}
Next, self-loops are removed from \(\hat{\mathbf{A}}^{l+1}\), as self-connections are already incorporated in the GCL through the inclusion of each node's features during the convolution operation (see Equation~\ref{eq:GCL}). 
The modified adjacency matrix, \(\check{\mathbf{A}}^{l+1}\), obtained by setting the diagonal elements of \(\hat{\mathbf{A}}^{l+1}\) to zero, is normalized as \(\tilde{\textbf{A}}^{l+1} = \check{\textbf{D}}^{-1/2} \check{\textbf{A}}^{l+1} \check{\textbf{D}}^{-1/2}\), where \(\check{\textbf{D}}\) represents the degree matrix of \(\check{\textbf{A}}^{l+1}\). 
Finally, the normalized adjacency matrix \(\tilde{\textbf{A}}^{l+1}\) is thresholded to generate a binarized representation, defined as \(\textbf{A}^{l+1} = \mathbb{I}(\tilde{\textbf{A}}^{l+1} \geq \tau)\), where \(\mathbb{I}(\cdot)\) is the indicator function that returns 1 if the condition is true and 0 otherwise, and \(\tau\) is the threshold value. This binarized matrix serves as the adjacency matrix for the next hierarchical level.

The encoder compresses the input graph until it is represented by a single node, which acts as a global descriptor, encapsulating the entire graph's structural and feature information. As a Variational Autoencoder (VAE), GAUDI maps the graph into a latent space, encoding it as a probabilistic multivariate Gaussian distribution. This distribution is parameterized by a mean vector \(\boldsymbol{\mu} \in \mathbb{R}^d\) and a standard deviation vector \(\boldsymbol{\sigma} \in \mathbb{R}^d\), where \(d\) is the dimensionality of the latent space. The latent variable \(\textbf{z} \in \mathbb{R}^d\) is sampled as \(\textbf{z} \sim \mathcal{N}(\boldsymbol{\mu}, \text{diag}(\boldsymbol{\sigma}^2))\) and the sampling process is performed using the reparameterization trick to ensure differentiability. Specifically, \(\textbf{z}\) is computed as:
\begin{equation} 
\textbf{z} = \boldsymbol{\mu} + \boldsymbol{\epsilon} \cdot \boldsymbol{\sigma},
\end{equation}
where \(\boldsymbol{\epsilon} \sim \mathcal{N}(0, \textbf{I})\) is sampled from a standard normal distribution. Furthermore, to ensure the latent space is structured and meaningful, a regularization term is added during training, aligning the learned distribution with the prior, \(\mathcal{N}(0, \textbf{I})\) (see Section \nameref{subsec:methods_training}). 

The graph is reconstructed from its latent representation through the decoder. The decoder, designed to mirror the encoder, consists of three blocks, each comprising a Graph Upsampling (GU) operation followed by a GCL (Fig.~\ref{fig:network}d). 
A key aspect of GAUDI's architecture is that the cluster assignment matrices and the binarized adjacency matrices at each level bypass the latent space and connect directly to the decoder using skip connections.
Each GU operation takes as input the node features from the previous level, $\mathbf{V}_{\text{dec}}^{l-1}$, along with the cluster assignment matrix, $\mathbf{S}^{l_{\text{enc}}}$, from the current level. The operation upsamples the node features as,
\begin{equation}
\mathbf{V}_{\text{up}}^{l} = \mathbf{S}^{l_{\text{enc}}} \mathbf{V}_{\text{dec}}^{l-1}.
\end{equation}
where, \( \mathbf{V}_{\text{dec}}^{l-1} = z\) in the first decoder block.
Subsequently, the upsampled node features undergo a graph convolution operation, as defined in Equation (3), using the binary adjacency matrix, \( \mathbf{A}\), from the corresponding encoder level \(l_{\text{enc}}\).

After the decoder, GAUDI reconstructs the input node and edge features of the graph using fully connected layers.
A three-layer fully connected neural network processes the decoded features to reconstruct the node features. The hidden layers employ ReLU activation, while the final layer has no activation function, enabling the network to autonomously determine the features to output.
The edge features, in turn, are reconstructed by first concatenating the node features of the connected nodes in the graph. These concatenated features are then passed through a two-layer, fully-connected neural network. The first layer uses ReLU activation, while the second has no activation function, ensuring the output dimensions align with those of the input edge features.

\subsection*{\label{subsec:methods_training}Details on the training of GAUDI}

GAUDI is trained to minimize the difference in node and edge features from the decoder output compared to the preprocessed graphs, using the mean absolute error. In addition to this loss, there is a weighted Kullback-Leibler divergence loss \cite{Kullback1951}, and a loss for the MinCut pooling \cite{Bianchi2020}. The total loss is a weighted sum of the losses, with the weights $\alpha$ for the node feature reconstruction loss, $\beta$ for the edge feature reconstruction loss, $\gamma$ for the Kullback-Leibler divergence, and $\delta$ for the MinCut pooling loss. The MinCut pooling loss is defined as
\begin{equation}
    \mathcal{L}_{\text{MinCut}} = \sum_{l \in \mathcal{P}} \biggl(- \frac{Tr((\mathbf{S}^l)^T \mathbf{A}^l \mathbf{S}^l)}{Tr((\mathbf{S}^l)^T \mathbf{D}^l \mathbf{S}^l)} + \left\lVert \frac{(\mathbf{S}^l)^T \mathbf{S}^l}{ \left\lVert (\mathbf{S}^l)^T \mathbf{S}^l \right\rVert _F} - \frac{\mathbf{I}_{N_v^{l+1}}}{\sqrt{N_v^{l+1}}} \right\rVert _F \biggr)
\end{equation}
with $\left\lVert . \right\rVert _F$ denoting the Frobenius norm, $\mathbf{D}^l$ denoting the degree matrix of $\mathbf{A}^l$, and $N_v^{l+1}$ denoting the number of nodes of the pooled graph. The set $\mathcal{P}$ indexes the encoder layers at which MinCut pooling is applied, including all encoder layers except the final one, where global pooling is performed.

We use the following parameters to obtain the described results. All of the fully connected layers of GAUDI have a dimension of 96, unless specified otherwise. The latent space dimension is set to 8. The graphs are pooled down to $N_v^{1} = 20$ nodes and $N_v^{2} = 5$ nodes in the blocks of the encoder. The threshold for the binarization of the adjacency matrices after the pooling are set to $\tau_1 = 1/19$ and $\tau_2 = 1/5$. The choice of values for the thresholds are motivated by the aim to keep at least one connection for each node of the pooled graph, while not having a fully connected graph. 

The network was trained for a varying number of epochs depending on the dataset, using different batch sizes and hyperparameters. The reconstruction loss for the node features was weighted by $\alpha$, while the reconstruction loss for the edge weights was weighted by $\gamma$. The Kullback-Leibler divergence was scaled by $\beta$, and the MinCut pooling loss was always weighted by 1. The ADAM optimizer \cite{kingma_adam_2017} was used with a learning rate $\zeta$.

Different hyperparameters were used for each dataset. For the Watts-Strogatz graphs, the model was trained for 5 epochs with loss function weights $\alpha = 0$, $\gamma = 10$, and $\beta = 1e^{-5}$, and a learning rate of $\zeta = 5e^{-5}$.
For the single molecule localization microscopy data, the model was trained for 25 epochs with $\alpha = 2$, $\gamma = 5$, and $\beta = 1e^{-7}$, and a learning rate of $\zeta = 1e^{-4}$. Similarly, for the Vicsek model simulations, GAUDI was trained for 20 epochs with $\alpha = 1$, $\gamma = 10$, and $\beta = 1e^{-3}$, and $\zeta = 1e^{-5}$, while for the brain connectivity graphs, training lasted 10 epochs with $\alpha = 2$, $\gamma = 5$, and $\beta = 1e^{-4}$, and $\zeta = 1e^{-4}$.

\subsection*{\label{subsec:methods_datasets}Details on the used samples and graphs}

\subsubsection*{Watts-Strogatz Small-World graphs}

Watts-Strogatz small-world graphs are composed by nodes that placed in a ring and being connected in a manner provided by the two parameters $C$, the number of neighboring nodes, and $p$, the rewiring probability \cite{Watts1998}. Here, we use 80 nodes for each graph. To begin with, each node is connected to it's $C$ nearest neighbors, where $C$ denotes an even number. Here, we use the values 2, 4, 6, and 8 for $C$. From this representation each edge is rewired once with the probability $p$, ranging from $0$ to $1$. This rewiring removes the existing edge, and connects the node to a randomly assigned node. This results in a regular graph for $p$ equal or close to zero, while an increasing $p$ leads to a more random graph (Fig.~\ref{fig:Watts-Strogatz}a).

When using GAUDI to analyze those Watts-Strogatz graphs the graphs are first normalized by centering the graphs around (0,0) and dividing the coordinates by the standard deviation of the coordinates of the graph. The node features of the graph are set to be the inverse distance of their 2-dimensional coordinate to the center of the graph, while the edge features are set to be the inverse length of the edges. The motivation behind using those node and edge features, instead of the actual coordinates of the nodes, is to make GAUDI focus on the global structure of the graph instead of the exact node positions.

\subsubsection*{Protein Assemblies from Single-Molecule Localization Microscopy Data}

Single-molecule localization microscopy (SMLM) is a form of stochastic super-resolution microscopy that enables imaging at resolutions beyond the diffraction limit. This technique works by precisely determining the spatial coordinates of individual fluorescent molecules. To achieve this high precision, it is crucial that the point spread functions of the fluorescent molecules do not overlap. This is accomplished by temporally separating the fluorescent emissions of the molecules. One method to achieve this temporal separation is to use fluorescent molecules that can photoswitch and capture multiple images over time, as in Stochastic Optical Reconstruction Microscopy (STORM)~\cite{rust_sub-diffraction-limit_2006, lelek2021}, which serves as the baseline for the creation of this dataset. This approach ensures that only a sparse subset of molecules emits light at any given moment, thereby allowing for the accurate localization of individual molecules in every frame. By combining multiple frames, a point cloud is obtained.

Inspired by experimental data of protein assemblies captured using STORM, we simulate 10,000 protein assemblies that either cover a spot-like area or form a ring shape with a 50\% probability. To mimic biological systems, we consider a hierarchical organization where proteins form small complexes that are, in turn, organized into spots or rings.

Each complex is composed of three proteins. The position of each protein relative to the complex center is determined by adding a normally distributed offset with a standard deviation of \qty{13}{\nano\meter}. The simulation incorporates incomplete labeling by assigning each protein a 50\% chance of being labeled. For each labeled protein, the number of localization events is randomly determined by a geometric distribution with a mean of eight localizations per protein. These localizations are displaced from the actual protein position by the addition of Gaussian noise with a standard deviation of \qty{20}{\nano\meter}, corresponding to the typical localization precision of STORM experiments.

For each sample simulation, a random number of complexes (between 6 and 15) is generated. If the complexes are arranged in a ring shape, the ring radius is randomly chosen between 50 and \qty{100}{\nano\meter}. Additionally, a random radial error is added to each complex’s radius, drawn from a normal distribution with a standard deviation between 0 and $\sqrt{2} \cdot \qty{20}{\nano\meter}$. If the complexes are arranged in a spot-like area, a radius between 50 and \qty{100}{\nano\meter} is randomly determined, and an additional displacement is added to the radius, drawn from a normal distribution with a standard deviation equal to $\sqrt{2}$ times the radius. In both configurations, the angular positions of the complexes are drawn from a uniform distribution over $[0, 2\pi)$ radians.

Lastly, a Poisson-distributed number of random noise points is added. These noise points are uniformly distributed within a square area centered on the sample, extending from -2 to +2 times the radius of the spot or ring.

To form a graph from a sample, the data points are treated as nodes, and each node is connected to its six most distant neighbors within the sample. The graphs are normalized by centering the coordinates of the nodes around $(0,0)$ and dividing the coordinates by the standard deviation of all coordinates in the graph. The node features are defined as the inverse distance from their 2-dimensional coordinates to the center of the graph, while the edge features are defined as the inverse length of the edge.

\subsubsection*{Vicsek model simulations}
In the Vicsek model, the point-like particles move at a constant speed and update their orientation by averaging the orientations of neighboring particles within the flocking radius $R_{\rm f}$, while incorporating random noise scaled by the noise level $\eta$ \cite{Vicsek1995}. By varying this radius and the size of the noise, the system can transition from a gas-like phase, where particles move almost independently, to a swarming phase, where particles self-organize into clusters. The positions of the particles are updated by moving them a step in their new direction. The simulation is conducted within a square area with periodic boundary conditions and the initial positions and orientations of the particles are randomly assigned.

In each simulation, we use $N = 100$ particles within a square with a side length equal to 100 units. The particles maintain a constant speed set to $v=1$ unit per time step, and the time step is defined as $\Delta t = 1$. The orientation of particle $i$ at time step $t+1$ is updated using the following equation: $\theta_{i,t+1} = \langle \theta_{j,t} \rangle + \eta \cdot W_{i,t} \cdot \Delta t$, where particles $j$ have a distance smaller than the flocking radius $R_{\rm f}$ to particle $i$ ($|\mathbf{r}_{j, t} - \mathbf{r}_{i, t}| < R_{\rm f}$), and $ W_{i,t}$ are random numbers from the uniform distribution $[-\frac{1}{2}, \frac{1}{2}]$.
Once the orientations have been updated, the positions are adjusted: $\mathbf{r}_{i,t+1} = \mathbf{r}_{i, t} + \mathbf{v} \cdot \Delta t $, with the velocity $\mathbf{v} = v \cdot (cos(\theta_{i,t}) \hat{x} + sin(\theta_{i,t}) \hat{y} )$

To ascertain the achievement of steady state in the systems, we compute two essential metrics: the global alignment coefficient $\Psi_t = \frac{1}{N} \big|\sum_{j=1}^{N} \frac{\mathbf{v}_{j,t}}{v}\big|$ and the global clustering coefficient $c_t = \frac{\text{count} \left\{ A_{i, t} < \pi R_{f}^{2} \right\} } {N}$, with $A_{i, t}$ being the area of the Voronoi polygon of particle $i$ at time step $t$.

We conduct 400 simulations, each with a flocking radius of either $R_{\rm f} = 1$ or $R_{\rm f} = 2$, and with a randomly assigned noise level $\eta$ between 0 and 1. To ensure the system has reached a steady state and to minimize the influence of initial conditions, each simulation runs for 8000 time steps and only the last 1000 time steps are analyzed. This number of time steps is motivated by the stabilization of the global alignment coefficient and the global clustering coefficient of the simulations. 

To construct a graph from a simulation time frame, particles are represented by nodes, each connected to its twelve nearest neighbors, taking periodic boundary conditions into account. The feature of the nodes is set to be the number of nodes that the node is connected to, while the edge feature is the inverse distance of the two nodes that are connected by the edge.

\subsubsection*{Brain connections}

The brain graphs analyzed here are derived from diffusion-weighted MRI (DWI) scans of 633 participants from the CamCAN dataset \cite{cam-can_cambridge_2014, taylor_cambridge_2017}. The anatomical connection strengths between the 120 regions of the automated anatomical atlas 2 (AAL2) \cite{rolls2015implementation} were extracted from these DWI scans. These connections represent the degree of integrity of the white matter fiber bundles between different brain regions, measured by fractional anisotropy. To focus on the most significant connections, a threshold was applied to retain only the top 20 percent of the connections with highest values. The coordinates of the nodes of the graphs are normalized by centering them around (0,0,0) and by dividing them by the standard deviation of the coordinates of the graph. The resulting graph features are the inverse distance of the nodes from the graph's center as node features and the inverse lengths of the edges as edge features.

\subsection*{\label{subsec:metrics}Metrics for performance evaluation}

To quantitatively evaluate the structure of the learned latent representations, we compute a set of metrics designed to assess both local clustering and the preservation of continuous parameters.

To evaluate whether continuous graph-generating parameters vary smoothly within discrete classes, we define a class-wise Isomap gradient continuity score. For each class, we first project the latent vectors onto their first two principal components (PC1 and PC2) and embed the projected points into a one-dimensional manifold using Isomap~\cite{Tenenbaum_global_2000} with $k=8$ nearest neighbors. This allows us to capture smooth variation even when the latent manifold is curved. We then compute the Spearman rank-order correlation coefficient~\cite{spearman_proof_1904, zwillinger_standard_2000} between the resulting Isomap coordinates and the continuous parameter of interest. Taking the absolute value gives the degree of monotonic variation within the class, with a value of 1 indicating perfect monotonicity. The final score is computed as the mean absolute Spearman correlation across all classes.

For the Watts–Strogatz graphs, we compute this score with respect to the rewiring probability $p$, using the number of initial neighbors $C$ to define the classes. For graphs generated by the Vicsek model, the classes are defined by the flocking radius $R_f$, and the continuous parameter is the noise level $\eta$. For the brain connectivity dataset, the continuous parameter is the participants’ age, and since no discrete classes are defined, the score is computed over the entire dataset as a single group. This metric is not computed for the SMLM dataset, which lacks an associated continuous parameter.

For methods such as MIAGAE with an 8-dimensional latent space, we observe that the Isomap gradient continuity score can be low even when the AUC or $R^2$ values are moderate. In these cases, the structure of the latent space projection may affect the outcome of the Isomap embedding. Specifically, for the brain connectivity dataset, the first two principal components of the MIAGAE-8 latent space form a nearly circular distribution, with most of the variation captured along the first principal component. Since Isomap extracts the dominant one-dimensional manifold within each class, the resulting embedding may not align with the direction of variation associated with the continuous parameter (in this case, age), which is primarily captured by the second principal component. This misalignment leads to a reduced absolute Spearman correlation and consequently a lower continuity score.

To assess the separability of graphs in latent space based on discrete labels, we use a linear support vector machine (SVM) and evaluate classification performance using the area under the receiver operating characteristic curve (AUC). For the SMLM dataset, the task is to distinguish ring-like from spot-like structures. For the brain connectivity data, participants are divided into two age groups based on the median age (55 years). The AUC is computed using both the first two principal components and the full eight-dimensional latent representations, where applicable.

For the brain connectivity data, we further assess how much of the variance in age is captured by the latent representation by computing the coefficient of determination ($R^2$)~\cite{wright1921correlation}. This is done by fitting a linear regression model with age as the target variable and either the first two principal components or all eight latent dimensions as input features.

\bibliography{apssamp}

\providecommand{\noopsort}[1]{}\providecommand{\singleletter}[1]{#1}%
\begin{thebibliography}{10}
\expandafter\ifx\csname url\endcsname\relax
  \def\url#1{\burl{#1}}\fi
\expandafter\ifx\csname urlprefix\endcsname\relax\def\urlprefix{URL }\fi
\providecommand{\bibinfo}[2]{#2}
\providecommand{\eprint}[2][]{\url{#2}}
\providecommand{\doi}[1]{\url{https://doi.org/#1}}
\bibcommenthead

\bibitem{Argun2021}
\bibinfo{author}{Argun, A.}, \bibinfo{author}{Callegari, A.} \& \bibinfo{author}{Volpe, G.}
\newblock \emph{\bibinfo{title}{Simulation of Complex Systems}}  (\bibinfo{publisher}{IOP Publishing}, \bibinfo{year}{2021}).
\newblock \urlprefix\url{https://doi.org/10.1088/978-0-7503-3843-1}.

\bibitem{jalving2019graph}
\bibinfo{author}{Jalving, J.}, \bibinfo{author}{Cao, Y.} \& \bibinfo{author}{Zavala, V.~M.}
\newblock \bibinfo{title}{Graph-based modeling and simulation of complex systems}.
\newblock \emph{\bibinfo{journal}{Computers \& Chemical Engineering}} \textbf{\bibinfo{volume}{125}}, \bibinfo{pages}{134--154} (\bibinfo{year}{2019}).
\newblock \urlprefix\url{https://www.sciencedirect.com/science/article/pii/S0098135418312687}.

\bibitem{holovatch_complex_2017}
\bibinfo{author}{Holovatch, Y.}, \bibinfo{author}{Kenna, R.} \& \bibinfo{author}{Thurner, S.}
\newblock \bibinfo{title}{Complex systems: physics beyond physics}.
\newblock \emph{\bibinfo{journal}{European Journal of Physics}} \textbf{\bibinfo{volume}{38}}, \bibinfo{pages}{023002} (\bibinfo{year}{2017}).
\newblock \urlprefix\url{https://iopscience.iop.org/article/10.1088/1361-6404/aa5a87}.

\bibitem{rathkopf_network_2018}
\bibinfo{author}{Rathkopf, C.}
\newblock \bibinfo{title}{Network representation and complex systems}.
\newblock \emph{\bibinfo{journal}{Synthese}} \textbf{\bibinfo{volume}{195}}, \bibinfo{pages}{55--78} (\bibinfo{year}{2018}).
\newblock \urlprefix\url{http://link.springer.com/10.1007/s11229-015-0726-0}.

\bibitem{Watts1998}
\bibinfo{author}{Watts, D.~J.} \& \bibinfo{author}{Strogatz, S.~H.}
\newblock \bibinfo{title}{Collective dynamics of `small-world' networks}.
\newblock \emph{\bibinfo{journal}{Nature}} \textbf{\bibinfo{volume}{393}}, \bibinfo{pages}{440--442} (\bibinfo{year}{1998}).
\newblock \urlprefix\url{https://doi.org/10.1038/30918}.

\bibitem{mijalkov_braph_2017}
\bibinfo{author}{Mijalkov, M.} \emph{et~al.}
\newblock \bibinfo{title}{{BRAPH}: {A} graph theory software for the analysis of brain connectivity}.
\newblock \emph{\bibinfo{journal}{PLOS ONE}} \textbf{\bibinfo{volume}{12}}, \bibinfo{pages}{1--23} (\bibinfo{year}{2017}).
\newblock \urlprefix\url{https://doi.org/10.1371/journal.pone.0178798}.
\newblock \bibinfo{note}{Publisher: Public Library of Science}.

\bibitem{bode2011impact}
\bibinfo{author}{Bode, N.~W.}, \bibinfo{author}{Wood, A.~J.} \& \bibinfo{author}{Franks, D.~W.}
\newblock \bibinfo{title}{The impact of social networks on animal collective motion}.
\newblock \emph{\bibinfo{journal}{Animal Behaviour}} \textbf{\bibinfo{volume}{82}}, \bibinfo{pages}{29--38} (\bibinfo{year}{2011}).
\newblock \urlprefix\url{https://www.sciencedirect.com/science/article/pii/S0003347211001539}.

\bibitem{battaglia2018relational}
\bibinfo{author}{Battaglia, P.~W.} \emph{et~al.}
\newblock \bibinfo{title}{Relational inductive biases, deep learning, and graph networks}.
\newblock \emph{\bibinfo{journal}{arXiv preprint arXiv:1806.01261}}  (\bibinfo{year}{2018}).
\newblock \urlprefix\url{https://arxiv.org/abs/1806.01261}.

\bibitem{velivckovic2023everything}
\bibinfo{author}{Veli{\v{c}}kovi{\'c}, P.}
\newblock \bibinfo{title}{Everything is connected: Graph neural networks}.
\newblock \emph{\bibinfo{journal}{Current Opinion in Structural Biology}} \textbf{\bibinfo{volume}{79}}, \bibinfo{pages}{102538} (\bibinfo{year}{2023}).
\newblock \urlprefix\url{https://www.sciencedirect.com/science/article/pii/S0959440X2300012X}.

\bibitem{ha_unraveling_2021}
\bibinfo{author}{Ha, S.} \& \bibinfo{author}{Jeong, H.}
\newblock \bibinfo{title}{Unraveling hidden interactions in complex systems with deep learning}.
\newblock \emph{\bibinfo{journal}{Scientific Reports}} \textbf{\bibinfo{volume}{11}}, \bibinfo{pages}{12804} (\bibinfo{year}{2021}).
\newblock \urlprefix\url{https://www.nature.com/articles/s41598-021-91878-w}.

\bibitem{wang_learning_2022}
\bibinfo{author}{Wang, R.}, \bibinfo{author}{Fang, F.}, \bibinfo{author}{Cui, J.} \& \bibinfo{author}{Zheng, W.}
\newblock \bibinfo{title}{Learning self-driven collective dynamics with graph networks}.
\newblock \emph{\bibinfo{journal}{Scientific Reports}} \textbf{\bibinfo{volume}{12}}, \bibinfo{pages}{500} (\bibinfo{year}{2022}).
\newblock \urlprefix\url{https://www.nature.com/articles/s41598-021-04456-5}.

\bibitem{pineda_geometric_2023}
\bibinfo{author}{Pineda, J.} \emph{et~al.}
\newblock \bibinfo{title}{Geometric deep learning reveals the spatiotemporal features of microscopic motion}.
\newblock \emph{\bibinfo{journal}{Nature Machine Intelligence}} \textbf{\bibinfo{volume}{5}}, \bibinfo{pages}{71--82} (\bibinfo{year}{2023}).
\newblock \urlprefix\url{https://doi.org/10.1038/s42256-022-00595-0}.

\bibitem{velivckovic2018deep}
\bibinfo{author}{Veličković, P.} \emph{et~al.}
\newblock \bibinfo{title}{Deep graph infomax}.
\newblock \emph{\bibinfo{journal}{arXiv preprint arXiv:1809.10341}}  (\bibinfo{year}{2018}).
\newblock \urlprefix\url{https://arxiv.org/abs/1809.10341}.

\bibitem{you_graph_2021}
\bibinfo{author}{You, Y.} \emph{et~al.}
\newblock \bibinfo{title}{Graph {Contrastive} {Learning} with {Augmentations}}.
\newblock \emph{\bibinfo{journal}{arXiv preprint arXiv:2010.13902}}  (\bibinfo{year}{2021}).
\newblock \urlprefix\url{http://arxiv.org/abs/2010.13902}.

\bibitem{ding2023eliciting}
\bibinfo{author}{Ding, K.}, \bibinfo{author}{Wang, Y.}, \bibinfo{author}{Yang, Y.} \& \bibinfo{author}{Liu, H.}
\newblock \emph{\bibinfo{title}{Eliciting structural and semantic global knowledge in unsupervised graph contrastive learning}}, Vol.~\bibinfo{volume}{37}, \bibinfo{pages}{7378--7386} (\bibinfo{publisher}{AAAI Press}, \bibinfo{year}{2023}).
\newblock \urlprefix\url{https://doi.org/10.1609/aaai.v37i6.25898}.

\bibitem{li2021prototypical}
\bibinfo{author}{Li, J.}, \bibinfo{author}{Zhou, P.}, \bibinfo{author}{Xiong, C.} \& \bibinfo{author}{Hoi, S.~C.}
\newblock \emph{\bibinfo{title}{Prototypical contrastive learning of unsupervised representations}}, \bibinfo{pages}{4--8} (\bibinfo{year}{2021}).

\bibitem{lelek2021}
\bibinfo{author}{Lelek, M.} \emph{et~al.}
\newblock \bibinfo{title}{Single-molecule localization microscopy}.
\newblock \emph{\bibinfo{journal}{Nature reviews methods primers}} \textbf{\bibinfo{volume}{1}}, \bibinfo{pages}{39} (\bibinfo{year}{2021}).
\newblock \urlprefix\url{https://doi.org/10.1038/s43586-021-00038-x}.

\bibitem{Vicsek1995}
\bibinfo{author}{Vicsek, T.}, \bibinfo{author}{Czir\'ok, A.}, \bibinfo{author}{Ben-Jacob, E.}, \bibinfo{author}{Cohen, I.} \& \bibinfo{author}{Shochet, O.}
\newblock \bibinfo{title}{Novel type of phase transition in a system of self-driven particles}.
\newblock \emph{\bibinfo{journal}{Phys. Rev. Lett.}} \textbf{\bibinfo{volume}{75}}, \bibinfo{pages}{1226--1229} (\bibinfo{year}{1995}).
\newblock \urlprefix\url{https://link.aps.org/doi/10.1103/PhysRevLett.75.1226}.

\bibitem{cam-can_cambridge_2014}
\bibinfo{author}{{Cam-{CAN}}} \emph{et~al.}
\newblock \bibinfo{title}{The {C}ambridge {C}entre for {A}geing and {N}euroscience ({C}am-{CAN}) study protocol: a cross-sectional, lifespan, multidisciplinary examination of healthy cognitive ageing}.
\newblock \emph{\bibinfo{journal}{BMC Neurology}} \textbf{\bibinfo{volume}{14}}, \bibinfo{pages}{204} (\bibinfo{year}{2014}).
\newblock \urlprefix\url{https://bmcneurol.biomedcentral.com/articles/10.1186/s12883-014-0204-1}.

\bibitem{ge2021graph}
\bibinfo{author}{Ge, Y.}, \bibinfo{author}{Pang, Y.}, \bibinfo{author}{Li, L.} \& \bibinfo{author}{Itti, L.}
\newblock \emph{\bibinfo{title}{Graph autoencoder for graph compression and representation learning}} (\bibinfo{year}{2021}).
\newblock \urlprefix\url{https://openreview.net/forum?id=Bo2LZfaVHNi}.
\newblock \bibinfo{note}{Neural Compression: From Information Theory to Applications--Workshop@ ICLR 2021}.

\bibitem{wang_molecular_2022}
\bibinfo{author}{Wang, Y.}, \bibinfo{author}{Wang, J.}, \bibinfo{author}{Cao, Z.} \& \bibinfo{author}{Barati~Farimani, A.}
\newblock \bibinfo{title}{Molecular contrastive learning of representations via graph neural networks}.
\newblock \emph{\bibinfo{journal}{Nature Machine Intelligence}} \textbf{\bibinfo{volume}{4}}, \bibinfo{pages}{279--287} (\bibinfo{year}{2022}).
\newblock \urlprefix\url{https://www.nature.com/articles/s42256-022-00447-x}.

\bibitem{li2023comprehensive}
\bibinfo{author}{Li, P.}, \bibinfo{author}{Pei, Y.} \& \bibinfo{author}{Li, J.}
\newblock \bibinfo{title}{A comprehensive survey on design and application of autoencoder in deep learning}.
\newblock \emph{\bibinfo{journal}{Applied Soft Computing}} \textbf{\bibinfo{volume}{138}}, \bibinfo{pages}{110176} (\bibinfo{year}{2023}).
\newblock \urlprefix\url{https://www.sciencedirect.com/science/article/pii/S1568494623001941}.

\bibitem{pan_adversarially_2018}
\bibinfo{author}{Pan, S.} \emph{et~al.}
\newblock \emph{\bibinfo{title}{Adversarially {Regularized} {Graph} {Autoencoder} for {Graph} {Embedding}}}, \bibinfo{pages}{2609--2615} (\bibinfo{publisher}{International Joint Conferences on Artificial Intelligence Organization}, \bibinfo{year}{2018}).
\newblock \urlprefix\url{https://www.ijcai.org/proceedings/2018/362}.

\bibitem{guo_multi-scale_2022}
\bibinfo{author}{Guo, Z.}, \bibinfo{author}{Wang, F.}, \bibinfo{author}{Yao, K.}, \bibinfo{author}{Liang, J.} \& \bibinfo{author}{Wang, Z.}
\newblock \emph{\bibinfo{title}{Multi-{Scale} {Variational} {Graph} {AutoEncoder} for {Link} {Prediction}}}, \bibinfo{pages}{334--342} (\bibinfo{publisher}{ACM}, \bibinfo{address}{Virtual Event AZ USA}, \bibinfo{year}{2022}).
\newblock \urlprefix\url{https://dl.acm.org/doi/10.1145/3488560.3498531}.

\bibitem{wang_attributed_2019}
\bibinfo{author}{Wang, C.} \emph{et~al.}
\newblock \bibinfo{title}{Attributed {Graph} {Clustering}: {A} {Deep} {Attentional} {Embedding} {Approach}}.
\newblock \emph{\bibinfo{journal}{arXiv preprint arXiv:1906.06532}}  (\bibinfo{year}{2019}).
\newblock \urlprefix\url{http://arxiv.org/abs/1906.06532}.

\bibitem{ronneberger2015u}
\bibinfo{author}{Ronneberger, O.}, \bibinfo{author}{Fischer, P.} \& \bibinfo{author}{Brox, T.}
\newblock \emph{\bibinfo{title}{U-net: Convolutional networks for biomedical image segmentation}}, \bibinfo{pages}{234--241} (\bibinfo{organization}{Springer}, \bibinfo{year}{2015}).
\newblock \urlprefix\url{https://doi.org/10.1007/978-3-319-24574-4_28}.

\bibitem{alon2020bottleneck}
\bibinfo{author}{Alon, U.} \& \bibinfo{author}{Yahav, E.}
\newblock \bibinfo{title}{On the bottleneck of graph neural networks and its practical implications}.
\newblock \emph{\bibinfo{journal}{arXiv preprint arXiv:2006.05205}}  (\bibinfo{year}{2020}).
\newblock \urlprefix\url{https://arxiv.org/abs/2006.05205}.

\bibitem{Bianchi2020}
\bibinfo{author}{Bianchi, F.~M.}, \bibinfo{author}{Grattarola, D.} \& \bibinfo{author}{Alippi, C.}
\newblock \bibinfo{title}{Spectral clustering with graph neural networks for graph pooling}.
\newblock \emph{\bibinfo{journal}{Proceedings of the 37th International Conference on Machine Learning, PMLR}} \textbf{\bibinfo{volume}{119}}, \bibinfo{pages}{874–883} (\bibinfo{year}{2020}).

\bibitem{grattarola_understanding_2024}
\bibinfo{author}{Grattarola, D.}, \bibinfo{author}{Zambon, D.}, \bibinfo{author}{Bianchi, F.~M.} \& \bibinfo{author}{Alippi, C.}
\newblock \bibinfo{title}{Understanding {Pooling} in {Graph} {Neural} {Networks}}.
\newblock \emph{\bibinfo{journal}{IEEE Transactions on Neural Networks and Learning Systems}} \textbf{\bibinfo{volume}{35}}, \bibinfo{pages}{2708--2718} (\bibinfo{year}{2024}).
\newblock \urlprefix\url{https://doi.org/10.1109/TNNLS.2022.3190922}.

\bibitem{hoelz2011}
\bibinfo{author}{Hoelz, A.}, \bibinfo{author}{Debler, E.~W.} \& \bibinfo{author}{Blobel, G.}
\newblock \bibinfo{title}{The structure of the nuclear pore complex}.
\newblock \emph{\bibinfo{journal}{Annual review of biochemistry}} \textbf{\bibinfo{volume}{80}}, \bibinfo{pages}{613--643} (\bibinfo{year}{2011}).

\bibitem{huotari2011}
\bibinfo{author}{Huotari, J.} \& \bibinfo{author}{Helenius, A.}
\newblock \bibinfo{title}{Endosome maturation}.
\newblock \emph{\bibinfo{journal}{The EMBO journal}} \textbf{\bibinfo{volume}{30}}, \bibinfo{pages}{3481--3500} (\bibinfo{year}{2011}).
\newblock \urlprefix\url{https://doi.org/10.1038/emboj.2011.286}.

\bibitem{mannella2006}
\bibinfo{author}{Mannella, C.~A.}
\newblock \bibinfo{title}{Structure and dynamics of the mitochondrial inner membrane cristae}.
\newblock \emph{\bibinfo{journal}{Biochimica et Biophysica Acta (BBA)-Molecular Cell Research}} \textbf{\bibinfo{volume}{1763}}, \bibinfo{pages}{542--548} (\bibinfo{year}{2006}).
\newblock \urlprefix\url{https://www.sciencedirect.com/science/article/pii/S0167488906000851}.

\bibitem{murphy2011}
\bibinfo{author}{Murphy, M.~P.}
\newblock \bibinfo{title}{Reactive oxygen species and mitochondrial function in life span control}.
\newblock \emph{\bibinfo{journal}{Free Radical Biology and Medicine}} \textbf{\bibinfo{volume}{51}}, \bibinfo{pages}{927--935} (\bibinfo{year}{2011}).

\bibitem{heider2016}
\bibinfo{author}{Heider, M.~R.} \& \bibinfo{author}{Munson, M.}
\newblock \bibinfo{title}{The exocyst complex: new insights into subunit localization and function}.
\newblock \emph{\bibinfo{journal}{Nature Reviews Molecular Cell Biology}} \textbf{\bibinfo{volume}{13}}, \bibinfo{pages}{377--384} (\bibinfo{year}{2016}).

\bibitem{Puig-Tinto2025continuum}
\bibinfo{author}{Puig-Tint{\'o}, M.} \emph{et~al.}
\newblock \bibinfo{title}{Continuum architecture dynamics of vesicle tethering in exocytosis}.
\newblock \emph{\bibinfo{journal}{bioRxiv}}  (\bibinfo{year}{2025}).
\newblock \urlprefix\url{https://www.biorxiv.org/content/early/2025/02/05/2025.02.05.635468}.

\bibitem{mostowy2012}
\bibinfo{author}{Mostowy, S.} \& \bibinfo{author}{Cossart, P.}
\newblock \bibinfo{title}{Septins: the fourth component of the cytoskeleton}.
\newblock \emph{\bibinfo{journal}{Nature Reviews Molecular Cell Biology}} \textbf{\bibinfo{volume}{13}}, \bibinfo{pages}{183--194} (\bibinfo{year}{2012}).
\newblock \urlprefix\url{https://doi.org/10.1038/nrm3284}.

\bibitem{taylor_cambridge_2017}
\bibinfo{author}{Taylor, J.~R.} \emph{et~al.}
\newblock \bibinfo{title}{The {C}ambridge {C}entre for {A}geing and {N}euroscience ({C}am-{CAN}) data repository: Structural and functional mri, meg, and cognitive data from a cross-sectional adult lifespan sample}.
\newblock \emph{\bibinfo{journal}{NeuroImage}} \textbf{\bibinfo{volume}{144}}, \bibinfo{pages}{262--269} (\bibinfo{year}{2017}).
\newblock \urlprefix\url{https://linkinghub.elsevier.com/retrieve/pii/S1053811915008150}.

\bibitem{rolls2015implementation}
\bibinfo{author}{Rolls, E.~T.}, \bibinfo{author}{Joliot, M.} \& \bibinfo{author}{Tzourio-Mazoyer, N.}
\newblock \bibinfo{title}{Implementation of a new parcellation of the orbitofrontal cortex in the automated anatomical labeling atlas}.
\newblock \emph{\bibinfo{journal}{Neuroimage}} \textbf{\bibinfo{volume}{122}}, \bibinfo{pages}{1--5} (\bibinfo{year}{2015}).
\newblock \urlprefix\url{https://doi.org/10.1016/j.neuroimage.2015.07.075}.

\bibitem{chang2025braph}
\bibinfo{author}{Chang, Y.-W.} \emph{et~al.}
\newblock \bibinfo{title}{Braph 2: A flexible, open-source, reproducible, community-oriented, easy-to-use framework for network analyses in neurosciences}.
\newblock \emph{\bibinfo{journal}{bioRxiv}} \bibinfo{pages}{2025.04.11.648455} (\bibinfo{year}{2025}).
\newblock \urlprefix\url{https://doi.org/10.1101/2025.04.11.648455}.

\bibitem{wright1921correlation}
\bibinfo{author}{Wright, S.}
\newblock \bibinfo{title}{Correlation and causation}.
\newblock \emph{\bibinfo{journal}{Journal of Agricultural Research}} \textbf{\bibinfo{volume}{XX, No.7}}, \bibinfo{pages}{557 -- 585} (\bibinfo{year}{1921}).

\bibitem{chicco2021coefficient}
\bibinfo{author}{Chicco, D.}, \bibinfo{author}{Warrens, M.~J.} \& \bibinfo{author}{Jurman, G.}
\newblock \bibinfo{title}{The coefficient of determination r-squared is more informative than {SMAPE}, {MAE}, {MAPE}, {MSE} and {RMSE} in regression analysis evaluation}.
\newblock \emph{\bibinfo{journal}{Peerj computer science}} \textbf{\bibinfo{volume}{7}}, \bibinfo{pages}{e623} (\bibinfo{year}{2021}).
\newblock \urlprefix\url{https://doi.org/10.7717/peerj-cs.623}.

\bibitem{Tenenbaum_global_2000}
\bibinfo{author}{Tenenbaum, J.~B.}, \bibinfo{author}{de~Silva, V.} \& \bibinfo{author}{Langford, J.~C.}
\newblock \bibinfo{title}{A global geometric framework for nonlinear dimensionality reduction}.
\newblock \emph{\bibinfo{journal}{Science}} \textbf{\bibinfo{volume}{290}}, \bibinfo{pages}{2319--2323} (\bibinfo{year}{2000}).
\newblock \urlprefix\url{https://www.science.org/doi/abs/10.1126/science.290.5500.2319}.

\bibitem{spearman_proof_1904}
\bibinfo{author}{Spearman, C.}
\newblock \bibinfo{title}{The proof and measurement of association between two things}.
\newblock \emph{\bibinfo{journal}{The American Journal of Psychology}} \textbf{\bibinfo{volume}{15}}, \bibinfo{pages}{72--101} (\bibinfo{year}{1904}).
\newblock \urlprefix\url{http://www.jstor.org/stable/1412159}.

\bibitem{zwillinger_standard_2000}
\bibinfo{author}{Zwillinger, D.} \& \bibinfo{author}{Kokoska, S.}
\newblock \emph{\bibinfo{title}{CRC standard probability and statistics tables and formulae}}  (\bibinfo{publisher}{Chapman \& Hall/CRC}, \bibinfo{year}{2000}).
\newblock \bibinfo{note}{Section 14.7}.

\bibitem{higgins2017beta}
\bibinfo{author}{Higgins, I.} \emph{et~al.}
\newblock \emph{\bibinfo{title}{beta-vae: Learning basic visual concepts with a constrained variational framework}} (\bibinfo{year}{2017}).
\newblock \urlprefix\url{https://openreview.net/forum?id=Sy2fzU9gl}.

\bibitem{fernandez2024learning}
\bibinfo{author}{Fern\'andez-Fern\'andez, G.}, \bibinfo{author}{Manzo, C.}, \bibinfo{author}{Lewenstein, M.}, \bibinfo{author}{Dauphin, A.} \& \bibinfo{author}{Mu\~noz Gil, G.}
\newblock \bibinfo{title}{Learning minimal representations of stochastic processes with variational autoencoders}.
\newblock \emph{\bibinfo{journal}{Physical Review E}} \textbf{\bibinfo{volume}{110}} (\bibinfo{year}{2024}).
\newblock \urlprefix\url{https://doi.org/10.1103/PhysRevE.110.L012102}.

\bibitem{Kullback1951}
\bibinfo{author}{Kullback, S.} \& \bibinfo{author}{Leibler, R.~A.}
\newblock \bibinfo{title}{On information and sufficiency}.
\newblock \emph{\bibinfo{journal}{The Annals of Mathematical Statistics}} \textbf{\bibinfo{volume}{22}}, \bibinfo{pages}{79--86} (\bibinfo{year}{1951}).
\newblock \urlprefix\url{http://www.jstor.org/stable/2236703}.

\bibitem{kingma_adam_2017}
\bibinfo{author}{Kingma, D.~P.} \& \bibinfo{author}{Ba, J.}
\newblock \bibinfo{title}{Adam: {A} {Method} for {Stochastic} {Optimization}}.
\newblock \emph{\bibinfo{journal}{arXiv preprint arXiv:1412.6980}}  (\bibinfo{year}{2017}).
\newblock \urlprefix\url{http://arxiv.org/abs/1412.6980}.

\bibitem{rust_sub-diffraction-limit_2006}
\bibinfo{author}{Rust, M.~J.}, \bibinfo{author}{Bates, M.} \& \bibinfo{author}{Zhuang, X.}
\newblock \bibinfo{title}{Sub-diffraction-limit imaging by stochastic optical reconstruction microscopy ({STORM})}.
\newblock \emph{\bibinfo{journal}{Nature Methods}} \textbf{\bibinfo{volume}{3}}, \bibinfo{pages}{793--796} (\bibinfo{year}{2006}).
\newblock \urlprefix\url{https://www.nature.com/articles/nmeth929}.

\end{thebibliography}
\end{document}